\documentclass{article}

\PassOptionsToPackage{square,numbers,compress}{natbib}

\usepackage[final]{neurips_2020}
\usepackage[utf8]{inputenc} 
\usepackage[T1]{fontenc}    
\usepackage{hyperref}       
\usepackage{url}            
\usepackage{graphicx}       
\usepackage{booktabs}       
\usepackage{amsfonts}       
\usepackage{nicefrac}       
\usepackage{microtype}      
\usepackage{authblk}        
\usepackage{amssymb}        
\usepackage{amsmath}        
\usepackage[inline]{enumitem}
\usepackage{caption}
\usepackage[super]{nth}
\newcommand\Loss{\mathcal{L}}
\newcommand\Tau{\mathcal{T}}
\usepackage{subcaption}
\usepackage[toc,page]{appendix}

\title{3D Self-Supervised Methods for Medical Imaging}

\author[1,*]{%
  Aiham Taleb
}
\author[1,$\dagger$]{%
  Winfried Loetzsch
}
\author[1,$\dagger$]{%
  Noel Danz
}
\author[1,*]{%
  Julius Severin
}
\author[1,$\dagger$]{%
  Thomas Gaertner
}
\author[1,*]{%
  Benjamin Bergner
}
\author[1,*]{%
  Christoph Lippert
}
\affil[1]{Digital Health \& Machine Learning\\
  Hasso-Plattner-Institute, Potsdam University\\
  Germany 
}
\affil[*]{\texttt {\{firstname.lastname\}@hpi.de}}
\affil[$\dagger$]{\texttt {\{firstname.lastname\}@student.hpi.uni-potsdam.de}}

\begin{document}

\maketitle

\begin{abstract}
  Self-supervised learning methods have witnessed a recent surge of interest after proving successful in multiple application fields. 
  In this work, we leverage these techniques, and we propose 3D versions for five different self-supervised methods, in the form of proxy tasks. Our methods facilitate neural network feature learning from \emph{unlabeled} 3D images, aiming to reduce the required cost for expert annotation. 
  The developed algorithms are 3D Contrastive Predictive Coding, 3D Rotation prediction, 3D Jigsaw puzzles, Relative 3D patch location, and 3D Exemplar networks. 
  Our experiments show that pretraining models with our 3D tasks yields more powerful semantic representations, and enables solving downstream tasks more accurately and efficiently, compared to training the models from scratch and to pretraining them on 2D slices. 
  We demonstrate the effectiveness of our methods on three downstream tasks from the medical imaging domain: i) Brain Tumor Segmentation from 3D MRI, ii) Pancreas Tumor Segmentation from 3D CT, and iii) Diabetic Retinopathy Detection from 2D Fundus images. In each task, we assess the gains in data-efficiency, performance, and speed of convergence. Interestingly, we also find gains when transferring the learned representations, by our methods, from a large unlabeled 3D corpus to a small downstream-specific dataset. 
  We achieve results competitive to state-of-the-art solutions at a fraction of the computational expense. 
  We publish our implementations\footnote{https://github.com/HealthML/self-supervised-3d-tasks} for the developed algorithms (both 3D and 2D versions) as an open-source library, in an effort to allow other researchers to apply and extend our methods on their datasets. 
\end{abstract}

\section{Introduction}
Due to technological advancements in 3D sensing, the need for machine learning-based algorithms that perform analysis tasks on 3D imaging data has grown rapidly in the past few years~\cite{review_DL_4_3D,review_DL_4_3D2,review_DL_4_3D3}. 3D imaging has numerous applications, such as in Robotic navigation, in CAD imaging, in Geology, and in Medical Imaging. While we focus on medical imaging as a test-bed for our proposed 3D algorithms in this work, we ensure their applicability to other 3D domains.
Medical imaging plays a vital role in patient healthcare, as it aids in disease prevention, early detection, diagnosis, and treatment. Yet efforts to utilize advancements in machine learning algorithms are often hampered by the sheer expense of the expert annotation required~\cite{annotate}. 
Generating expert annotations of 3D medical images at scale is non-trivial, expensive, and time-consuming. 
Another related challenge in medical imaging is the relatively small sample sizes. This becomes more obvious when studying a particular disease, for instance. Also, gaining access to large-scale datasets is often difficult due to privacy concerns. Hence, scarcity of data and annotations are some of the main constraints for machine learning applications in medical imaging. 

Several efforts have attempted to address these challenges, as they are common to other application fields of deep learning. A widely used technique is transfer learning, which aims to reuse the features of already trained neural networks on different, but related, target tasks. A common example is adapting the features from networks trained on ImageNet, which can be reused for other visual tasks, e.g. semantic segmentation.
To some extent, transfer learning has made it easier to solve tasks with limited number of samples. However, as mentioned before, the medical domain is supervision-starved. Despite attempts to leverage ImageNet~\cite{imagenet_cvpr09} features in the medical context~\cite{xray_imagenet,xray_imagenet2,retinopathy_imagenet,retinopathy_imagenet2}, the difference in the distributions of natural and medical images is significant, i.e. generalizing across these domains is questionable and can suffer from dataset bias~\cite{dataset_bias}. Recent analysis~\cite{Transfusion} has also found that such transfer learning offers limited performance gains, relative to the computational costs it incurs. 
Consequently, it is necessary to find better solutions for the aforementioned challenges.

A viable alternative is to employ self-supervised (unsupervised) methods, which proved successful in multiple domains recently. In these approaches, the supervisory signals are derived from the data. In general, we withhold some part of the data, and train the network to predict it. This prediction task defines a proxy loss, which encourages the model to learn semantic representations about the concepts in the data. Subsequently, this facilitates data-efficient fine-tuning on supervised downstream tasks, reducing significantly the burden of manual annotation. 
Despite the surge of interest in the machine learning community in self-supervised methods, only little work has been done to adopt these methods in the medical imaging domain. We believe that self-supervised learning is directly applicable in the medical context, and can offer cheaper solutions for the challenges faced by conventional supervised methods. Unlabelled medical images carry valuable information about organ structures, and self-supervision enables the models to derive notions about these structures with no additional annotation cost. 

A particular aspect of most medical images, which received little attention by previous self-supervised methods, is their 3D nature~\cite{medical_imaging}. The common paradigm is to cast 3D imaging tasks in 2D, by extracting slices along an arbitrary axis, e.g. the axial dimension. However, such tasks can substantially benefit from the full 3D spatial context, thus capturing rich anatomical information. We believe that relying on the 2D context to derive data representations from 3D images, in general, is a suboptimal solution, which compromises the performance on downstream tasks. 

\textbf{Our contributions.} As a result, in this work, we propose five self-supervised tasks that utilize the full 3D spatial context, aiming to better adopt self-supervision in 3D imaging.
The proposed tasks are: 3D Contrastive Predictive Coding, 3D Rotation prediction, 3D Jigsaw puzzles, Relative 3D patch location, and 3D Exemplar networks. These algorithms are inspired by their successful 2D counterparts, and to the best of our knowledge, most of these methods have never been extended to the 3D context, let alone applied to the medical domain. Several computational and methodological challenges arise when designing self-supervised tasks in 3D, due to the increased data dimensionality, which we address in our methods to ensure their efficiency. 
We perform extensive experiments using four datasets in three different downstream tasks, and we show that our 3D tasks result in rich data representations that improve data-efficiency and performance on three different downstream tasks. 
Finally, we publish the implementations of our 3D tasks, and also of their 2D versions, in order to allow other researchers to evaluate these methods on other imaging datasets. 

\section{Related work}
In general, unsupervised representation learning can be formulated as learning an embedding space, in which data samples that are semantically similar are closer, and those that are different are far apart. The self-supervised family constructs such a representation space by creating a supervised proxy task from the data itself. Then, the embeddings that solve the proxy task will also be useful for other real-world downstream tasks. Several methods in this line of research have been developed recently, and they found applications in numerous fields~\cite{survey_self_supervised}. In this work, we focus on methods that operate on images only. 

Self-supervised methods differ in their core building block, i.e. the proxy task used to learn representations from unlabelled input data. 
A commonly used supervision source for proxy tasks is the spatial context from images, which was first inspired by the skip-gram Word2Vec~\cite{w2v} algorithm. This idea was generalized to images in~\cite{context_prediction}, in which a visual representation is learned by predicting the position of an image patch relative to another. A similar work extended this patch-based approach to solve Jigsaw Puzzles~\cite{jig}. Other works have used different supervision sources, such as image colors~\cite{color}, clustering~\cite{deep_cluster}, image rotation prediction~\cite{rotations}, object saliency~\cite{ssl_salient_beauty}, and image reconstruction~\cite{context_encoders}. In recent works, Contrastive Predictive Coding (CPC) approaches~\cite{CPC1,CPC2} advanced the results of self-supervised methods on multiple imaging benchmarks~\cite{chen2020simple,momentum_contrast}. These methods utilize the idea of contrastive learning in the latent space, similar to Noise Contrastive Estimation~\cite{NCE}. In 2D images, the model has to predict the latent representation for next (adjacent) image patches. Our work follows this line of research in the above works, however, our methods utilize the full 3D context.

While videos are rich with more types of supervisory signals~\cite{videos1,videos2,videos3,videos4,videos5}, we discuss here a subset of these works that utilize 3D-CNNs to process input videos. In this context, 3D-CNNs are employed to simultaneously extract spatial features from each frame, and temporal features across multiple frames, which are typically stacked along the \nth{3} (depth) dimension. The idea of exploiting 3D convolutions for videos was proposed in~\cite{3DCNN_video} for human action recognition, and was later extended to other applications~\cite{survey_self_supervised}. In self-supervised learning, however, the number of pretext tasks that exploit this technique is limited. Kim~\emph{et al.}~\cite{videos_3D_cubic_puzzles} proposed a task that extracts cubic puzzles of $2\times2\times1$, meaning that the \nth{3} dimension is not actually utilized in puzzle creation. Jing~\emph{et al.}~\cite{videos_3D_rot} extended the rotation prediction task~\cite{rotations} to videos, by simply stacking video frames along the depth dimension, however, this dimension is not employed in the design of their task as only spatial rotations are considered. Han \emph{et al.} proposed a dense encoding of spatio-temporal frame blocks to predict future scene representations recurrently, in conjunction with a curriculum training scheme to extend the predicted future. Similarly, the depth dimension is not employed in this task.
On the other hand, in our more general versions of 3D Jigsaw puzzles and 3D Rotation prediction, respectively, we exploit the depth (\nth{3}) dimension in the design of our tasks. For instance, we solve larger 3D puzzles up to $3\times3\times3$, and we also predict more rotations along all axes in the 3D space. Futhermore, in our 3D Contrastive Predictive Coding task, we predict patch representations along all 3 dimensions, scanning input volumes in a manner that resembles a pyramid. 
In general, we believe the different nature of the data, 3D volumetric scans vs. stacked video frames, influences the design of proxy tasks, i.e. the depth dimension has an actual semantic meaning in volumetric scans. Hence, we consider the whole 3D context when designing all of our methods, aiming to learn valuable anatomical information from unlabeled 3D volumetric scans. 

In the medical context, self-supervision has found use-cases in diverse applications such as depth estimation in monocular endoscopy~\cite{depth}, robotic surgery~\cite{surgery}, medical image registration~\cite{register}, body part recognition~\cite{body}, in disc degeneration using spinal MRIs~\cite{disc}, in cardiac image segmentation~\cite{cardiac_self_supervised}, body part regression for slice ordering~\cite{Yan2018DeepLG}, and medical instrument segmentation~\cite{endoscopic_videos}. Spitzer \emph{et al.}~\cite{Cytoarchitectonic_segmentation} sample 2D patches from a 3D brain, and predict the distance between these patches as a supervision signal. Tajbakhsh \emph{et al.}~\cite{orientation_prediction_tajbakhsh} use orientation prediction from medical images as a proxy task.
There are multiple other examples of self-supervised methods for medical imaging, such as~\cite{image_context_medical,ultrasound_video,multimodal_puzzles,image_context_medical2,chaitanya2020contrastive}.
While these attempts are a step forward for self-supervised learning in medical imaging, they have some limitations. First, as opposed to our work, many of these works make assumptions about input data, resulting in engineered solutions that hardly generalize to other target tasks. Second, none of the above works capture the complete spatial context available in 3-dimensional scans, i.e. they only operate on 2D/2.5D spatial context. 
In a more related work, Zhou \emph{et al.}~\cite{ssl_models_genesis} extended image reconstruction techniques from 2D to 3D, and implemented multiple self-supervised tasks based on image-reconstruction. 
Zhuang \emph{et al.}~\cite{rubik} and Zhu \emph{et al.}~\cite{rubik2} developed a proxy task that solves small 3D jigsaw puzzles. Their proposed puzzles were only limited to $2\times2\times2$ of puzzle complexity. Our version of 3D Jigsaw puzzles is able to efficiently solve larger puzzles, e.g. $3\times3\times3$, and outperforms their method's results on the downstream task of Brain tumor segmentation. 
In this paper, we continue this line of work, and develop five different algorithms for 3D data, whose nature and performance can accommodate more types of target medical applications. 

\section{Self-Supervised Methods}
In this section, we discuss the formulations of our 3D self-supervised pretext tasks, all of which learn data representations from unlabeled samples (3D images), hence requiring no manual annotation effort in the self-supervised pretraining stage. Each task results in a pretrained encoder model $g_{enc}$ that can be fine-tuned in various downstream tasks, subsequently. 

\begin{figure}[htb]
\centering
\includegraphics[width=\textwidth]{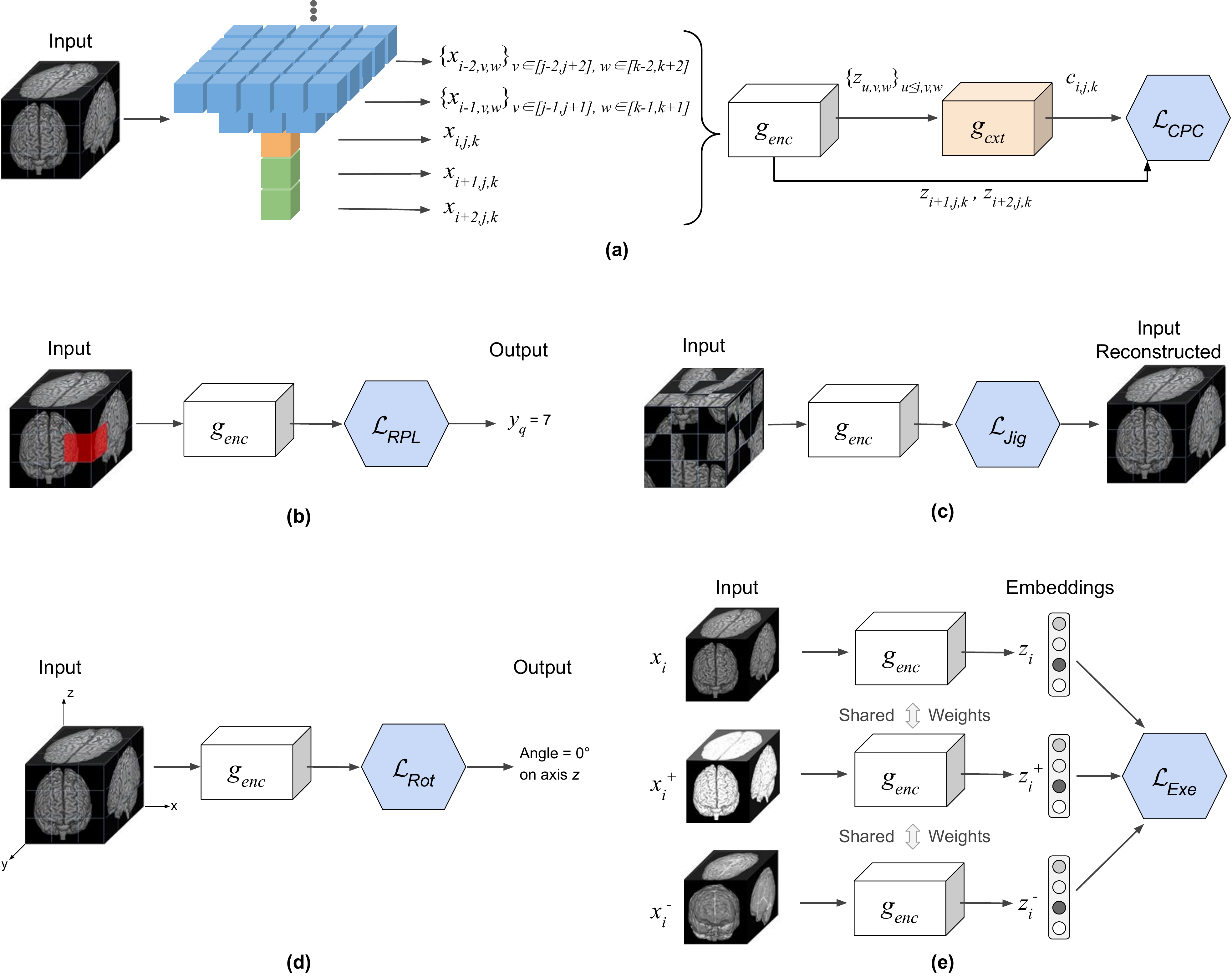}
\caption{\textbf{(a)} 3D-CPC: each input image is split into 3D patches, and the latent representations $z_{i+1,j,k}, z_{i+2,j,k}$ of next patches $x_{i+1,j,k}, x_{i+2,j,k}$ (shown in green) are predicted using the context vector $c_{i,j,k}$. The considered context is the current patch $x_{i,j,k}$ (shown in orange), plus the above patches that form an inverted pyramid (shown in blue). 
\textbf{(b)} 3D-RPL: assuming a 3D grid of 27 patches ($3\times3\times3$), the model is trained to predict the location $y_q$ of the query patch $x_q$ (shown in red), relative to the central patch $x_c$ (whose location is 13). 
\textbf{(c)} 3D-Jig: by predicting the permutation applied to the 3D image when creating a $3\times3\times3$ puzzle, we are able to reconstruct the scrambled input.
\textbf{(d)} 3D-Rot: the network is trained to predict the rotation degree (out of the 10 possible degrees) applied on input scans. 
\textbf{(e)} 3D-Exe: the network is trained with a triplet loss, which drives positive samples closer in the embedding space ($x_i^{+}$ to $x_i$), and the negative samples ($x_i^{-}$) farther apart. }
\label{fig_big}
\end{figure}

\subsection{3D Contrastive Predictive Coding (3D-CPC)} \label{cpc}
Following the contrastive learning idea, first proposed in~\cite{NCE}, this universal unsupervised technique predicts the latent space for future (next or adjacent) samples. Recently, CPC found success in multiple application fields, e.g. its 1D version in audio signals~\cite{CPC1}, and its 2D versions in images~\cite{CPC1,CPC2}, and was able to bridge the gap between unsupervised and fully-supervised methods~\cite{chen2020simple}. 
Our proposed CPC version generalizes this technique to 3D inputs, and defines a proxy task by cropping equally-sized and overlapping 3D patches from each input scan.
Then, the encoder model $g_{enc}$ maps each input patch $x_{i,j,k}$ to its latent representation $z_{i,j,k} = g_{enc}(x_{i,j,k})$.
Next, another model called the context network $g_{cxt}$ is used to summarize the latent vectors of the patches in the context of $x_{i,j,k}$, and produce its context vector $c_{i,j,k} = g_{cxt}(\{z_{u,v,w}\}_{u \leq i,v,w})$, where $\{z\}$ denotes a set of latent vectors.
Finally, because $c_{i,j,k}$ captures the high level content of the context that corresponds to $x_{i,j,k}$, it allows for predicting the latent representations of next (adjacent) patches $z_{i+l,j,k}$, where $l \geq 0$. This prediction task is cast as an $N$-way classification problem by utilizing the InfoNCE loss~\cite{CPC1}, which takes its name from its ability to maximize the mutual information between $c_{i,j,k}$ and $z_{i+l,j,k}$.
Here, the classes are the latent representations $\{z\}$ of the patches, among which is one \emph{positive} representation, and the rest $N-1$ are \emph{negative}. Formally, the CPC loss can be written as follows:
\begin{equation}
\begin{split}
    \Loss_{CPC} &= - \sum_{i,j,k,l} \log p(z_{i+l,j,k} \mid \hat{z}_{i+l,j,k}, \{ z_n \})  \\
    &= - \sum_{i,j,k,l} \log \frac{\exp(\hat{z}_{i+l,j,k} z_{i+l,j,k})}{\exp(\hat{z}_{i+l,j,k} z_{i+l,j,k}) + \exp(\sum_{n}\hat{z}_{i+l,j,k} z_n)}
\end{split}
\end{equation}
This loss corresponds to the categorical cross-entropy loss, which trains the model to recognize the correct representation $z_{i+l,j,k}$ among the list of negative representations $\{ z_n \}$. These negative samples (3D patches) are chosen randomly from other locations in the input image. In practice, similar to the original NCE~\cite{NCE}, this task is solved as a binary pairwise classification task.

It is noteworthy that the proposed 3D-CPC task, illustrated in Fig.~\ref{fig_big} (a), allows employing any network architecture in the encoder $g_{enc}$ and the context $g_{cxt}$ networks. In our experiments, we follow~\cite{CPC1} in using an autoregressive network using GRUs~\cite{GRU} for the context network $g_{cxt}$, however, masked convolutions can be a valid alternative~\cite{pixel_cnn}. In terms of what the 3D context of each patch $x_{i,j,k}$ includes, we follow the idea of an inverted pyramid neighborhood, which is inspired from~\cite{pyramid_LSTM,pixel_rnn}. This context is chosen based on a tradeoff between computational cost and performance. Too large contexts (e.g. full surrounding of a patch) incur prohibitive computations and memory use. The inverted-pyramid context was an optimal tradeoff. 

\subsection{Relative 3D patch location (3D-RPL)} \label{rpl}
In this task, the spatial context in images is leveraged as a rich source of supervision, in order to learn semantic representations of the data. First proposed by Doersch \emph{et al.}~\cite{context_prediction} for 2D images, this task inspired several works in self-supervision. 
In our 3D version, shown in Fig.~\ref{fig_big} (b), we leverage the full 3D spatial context in the design of our task. 
From each input 3D image, a 3D grid of $N$ non-overlapping patches $\{x_i\}_{i \in \{1,..,N\}}$ is sampled at random locations.
Then, the patch $x_c$ in the center of the grid is used as a reference, and a query patch $x_q$ is selected from the surrounding $N-1$ patches. 
Next, the location of $x_q$ relative to $x_c$ is used as the positive label $y_q$. This casts the task as an $N-1$-way classification problem, in which the locations of the remaining grid patches are used as the negative samples $\{y_n\}$. Formally, the cross-entropy loss in this task is written as:
\begin{equation}
    \Loss_{RPL} = - \sum_{k=1}^{K} \log p(y_q \mid \hat{y}_q, \{ y_n \}) 
\end{equation}
Where $K$ is the number of queries extracted from all samples.
In order to prevent the model from solving this task quickly by finding shortcut solutions, e.g. edge continuity, we follow~\cite{context_prediction} in leaving random gaps (jitter) between neighboring 3D patches. More details in Appendix.

\subsection{3D Jigsaw puzzle Solving (3D-Jig)} \label{jig}
Deriving a Jigsaw puzzle grid from an input image, be it in 2D or 3D, and solving it can be viewed as an extension to the above patch-based RPL task. 
In our 3D Jigsaw puzzle task, which is inspired by its 2D counterpart~\cite{jig} and illustrated in Fig.~\ref{fig_big} (c), the puzzles are formed by sampling an $n \times n \times n$ grid of 3D patches. 
Then, these patches are shuffled according to an arbitrary permutation, selected from a set of predefined permutations. This set of permutations with size $P$ is chosen out of the $n^3!$ possible permutations, by following the Hamming distance based algorithm in~\cite{jig} (details in Appendix), and each permutation is assigned an index $y_p \in \{1,..,P\}$. 
Therefore, the problem is cast as a $P$-way classification task, i.e., the model is trained to simply recognize the applied permutation index $p$, allowing us to solve the 3D puzzles in an efficient manner. Formally, we minimize the cross-entropy loss of $\Loss_{Jig} (y_{p}^{k},\hat{y}_{p}^{k})$, where $k \in \{1,..,K\}$ is an arbitrary 3D puzzle from the list of extracted $K$ puzzles.
Similar to 3D-RPL, we use the trick of adding random jitter in 3D-Jig.

\subsection{3D Rotation prediction (3D-Rot)} \label{rot}
Originally proposed by Gidaris \emph{et al.}~\cite{rotations}, the rotation prediction task encourages the model to learn visual representations by simply predicting the angle by which the input image is rotated. The intuition behind this task is that for a model to successfully predict the angle of rotation, it needs to capture sufficient semantic information about the object in the input image. 
In our 3D Rotation prediction task, 3D input images are rotated randomly by a random degree $r \in \{1,..,R\}$ out of the $R$ considered degrees. In this task, for simplicity, we consider the multiples of 90 degrees (0$^{\circ}$, 90$^{\circ}$, 180$^{\circ}$, 270$^{\circ}$, along each axis of the 3D coordinate system $(x, y, z)$. 
There are 4 possible rotations \emph{per axis}, amounting to 12 possible rotations. However, rotating input scans by 0$^{\circ}$ along the 3 axes will produce 3 identical versions of the original scan, hence, we consider 10 rotation degrees instead. Therefore, in this setting, this proxy task can be solved as a 10-way classification problem. Then, the model is tasked to predict the rotation degree (class), as shown in Fig.~\ref{fig_big} (d). Formally, we minimize the cross-entropy loss $\Loss_{Rot} (r^{k},\hat{r}^{k})$, where $k \in \{1,..,K\}$ is an arbitrary rotated 3D image from the list of $K$ rotated images. It is noteworthy that we create multiple rotated versions for each 3D image.

\subsection{3D Exemplar networks (3D-Exe)} \label{exe}
The task of Exemplar networks, proposed by Dosovitskiy \emph{et al.}~\cite{exemplar}, is one of the earliest methods in the self-supervised family. To derive supervision labels, it relies on image augmentation techniques, i.e. transformations. Assuming a training set $X = \{x_1,...x_N\}$, and a set of $K$ image transformations $\Tau = \{T_1,..T_K\}$, a new surrogate class $S_{x_i}$ is created by transforming each training sample $x_i \in X$, where $S_{x_i}=\Tau x_i = \{Tx_i \mid T \in \Tau\}$. Therefore, the task is cast as a regular classification task with a cross-entropy loss.
However, this classification task becomes prohibitively expensive as the dataset size grows larger, as the number of classes grows accordingly.
Thus, in our proposed 3D version of Exemplar networks, shown in Fig.~\ref{fig_big} (e), we employ a different mechanism that relies on the triplet loss instead~\cite{exemplar_triplet}. 
Formally, assuming $x_i$ is a random training sample and $z_i$ is its corresponding embedding vector, $x_i^{+}$ is a transformed version of $x_i$ (seen as a positive example) with an embedding $z_i^{+}$, and $x_i^{-}$ is a different sample from the dataset (seen as negative) with an embedding $z_i^{-}$. The triplet loss is written as follows:
\begin{equation}
    \Loss_{Exe} = \frac{1}{N_T} \sum_{i=1}^{N_T} \max \{0, D(z_i,z_i^{+}) - D(z_i,z_i^{-}) + \alpha\}
\end{equation}
where $D(.)$ is a pairwise distance function, for which we use the $L_2$ distance, following~\cite{FaceNet_triplet}. $\alpha$ is a margin (gap) that is enforced between positive and negative pairs, which we set to $1$. The triplet loss enforces $D(z_i,z_i^{-}) > D(z_i,z_i^{+})$, i.e. the transformed versions of the same sample (positive samples) to come closer to each other in the learned embedding space, and farther away from other (negative) samples. Replacing the triplet loss with a contrastive loss~\cite{NCE} is possible in this method, and has been found to improve learned representations from natural images~\cite{chen2020simple}. In addition, the learned representations by Exemplar can be affected by the negatives sampling strategy. The simple option is to sample from within the same batch, however, it is also possible to sample from the whole dataset. The latter choice is computationally more expensive, but is expected to improve the learned representations, as it makes the task harder.
It is noteworthy that we apply the following 3D transformations: random flipping along an arbitrary axis, random rotation along an arbitrary axis, random brightness and contrast, and random zooming.

\section{Experimental Results}
In this section, we present the evaluation results of our methods, which we assess the quality of their learned representations by fine-tuning them on three downstream tasks. In each task, we analyze the obtained gains in data-efficiency, performance, and speed of convergence. In addition, each task aims to demonstrate a certain use-case for our methods. We follow the commonly used evaluation protocols for self-supervised methods in each of these tasks.
The chosen tasks are: 
\begin{itemize}
    \item Brain Tumor Segmentation from 3D MRI (Subsection~\ref{brain}): in which we study the possibility for transfer learning from a different unlabeled 3D corpus, following~\cite{scaling_self_supervised}.
    \item Pancreas Tumor Segmentation from 3D CT (Subsection~\ref{pancreas}): to demonstrate how to use the same unlabeled dataset, following the data-efficient evaluation protocol in~\cite{CPC2}. 
    \item Diabetic Retinopathy Detection from 2D Fundus Images (Subsection~\ref{retino}): to showcase our implementations for the 2D versions of our methods, following~\cite{CPC2}. Here, we also evaluate pretraining on a different large corpus, then fine-tuning on the downstream dataset.
\end{itemize}
We provide additional details about architectures, training procedures, the effect of augmentation in Exemplar, and how we initialize decoders for segmentation tasks in the Appendix.

\subsection{Brain Tumor Segmentation Results} \label{brain} 
In this task, we evaluate our methods by fine-tuning the learned representations on the Multimodal Brain Tumor Segmentation (BraTS) 2018~\cite{brats1,brats2} benchmark. Before that, we pretrain our models on brain MRI data from the UK Biobank~\cite{ukbio} (UKB) corpus, which contains roughly $22K$ 3D scans. Due to this large number of unlabeled scans, UKB is suitable for unsupervised pretraining. The BraTS dataset contains annotated MRI scans for $285$ training and $66$ validation cases. We fine-tune on BraTS' training set, and evaluate on its validation set. Following the official BraTS challenge, we report Dice scores for the Whole Tumor (WT), Tumor Core (TC), and Enhanced Tumor (ET) tasks. The Dice score (F1-Score) is twice the area of overlap between two segmentation masks divided by the total number of pixels in both. 
In order to assess the quality of the learned representations by our 3D proxy tasks, we compare to the following baselines:
\begin{itemize}
    \item Training from scratch: the first sensible baseline for any self-supervised method, in general, is the same model trained on the downstream task when initialized from random weights. Comparing to this baseline provides insights about the benefits of self-supervised pretraining.
    \item Training on 2D slices: this baseline aims to quantitatively show how our proposal to operate on the 3D context benefits the learned representations, compared to 2D methods. 
    \item Supervised pretraining: this baseline uses automatic segmentation labels from FSL-FAST~\cite{fsl_fast}, which include masks for three brain tissues. 
    \item Baselines from the BraTS challenge: we compare to the methods~\cite{isensee,Popli,Baid,Chandra}, which all use a single model with an architecture similar to ours, i.e. 3D U-Net~\cite{UNET}.
\end{itemize}

\begin{figure}[!htb]
\minipage{0.45\textwidth}
  \small
  \includegraphics[width=\linewidth]{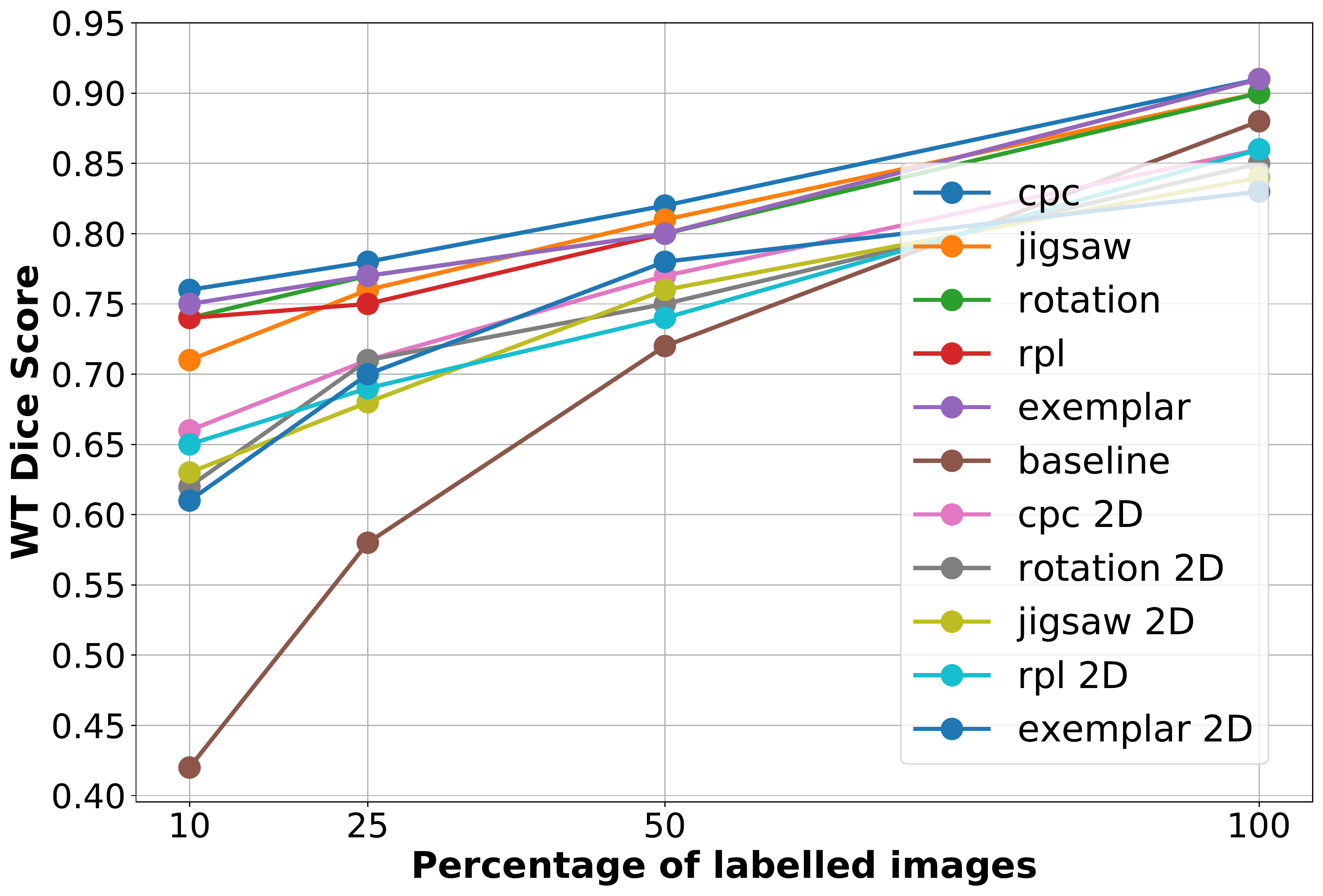}
  \caption{Data-efficient segmentation results in BraTS. With less labeled data, the supervised baseline (brown) fails to generalize, as opposed to our methods. Also, the proposed 3D methods outperform all 2D counterparts.}
  \label{plot_brats}
\endminipage\hfill
\minipage{0.45\textwidth}
    \centering          \small
    \captionof{table}{BraTS segmentation results} 
    \label{table_brain}
    \begin{tabular}[b]{ c c c c } \toprule
    Model & ET & WT & TC  \\
    \hline
    3D-From scratch                      & 76.38 & 87.82 & 83.11 \\
    3D Supervised                        & 78.88 & 90.11 & 84.92 \\
    \hline
    2D-CPC                               & 76.60 & 86.27 & 82.41 \\
    2D-RPL                               & 77.53 & 87.91 & 82.56 \\ 
    2D-Jigsaw                            & 76.12 & 86.28 & 83.26 \\ 
    2D-Rotation                          & 76.60 & 88.78 & 82.41 \\
    2D-Exemplar                          & 75.22 & 84.82 & 81.87 \\ 
    \hline 
    Popli \emph{et al.}~\cite{Popli}     & 74.39 & 89.41 & 82.48 \\
    Baid \emph{et al.}~\cite{Baid}       & 74.80 & 87.80 & 82.66 \\
    Chandra \emph{et al.}~\cite{Chandra} & 74.06 & 87.19 & 79.89 \\
    Isensee \emph{et al.}~\cite{isensee} & 80.36 & \textbf{90.80} & 84.32 \\ 
    \hline 
    3D-CPC                               & 80.83 & 89.88 & 85.11 \\
    3D-RPL                               & \textbf{81.28} & 90.71 & \textbf{86.12} \\
    3D-Jigsaw                            & 79.66 & 89.20 & 82.52 \\
    3D-Rotation                          & 80.21 & 89.63 & 84.75 \\
    3D-Exemplar                          & 79.46 & \textbf{90.80} & 83.87 \\
    \hline
    \end{tabular}
\endminipage\hfill
\end{figure}

\textbf{Discussion.} 
We first assess the gains in data-efficiency in this task. To quantify these gains, we measure the segmentation performance at different sample sizes. We randomly select subsets of patients at 10\%, 25\%, 50\%, and 100\% of the full dataset size, and we fine-tune our models on these subsets. Here, we compare to the baselines listed above. As shown in Fig.~\ref{plot_brats}, our 3D methods outperform the baseline model trained from scratch by a large margin when using few training samples, and behaves similarly as the number of labeled samples increases. The low-data regime case at 5\% suggests the potential for generic unsupervised features, and highlights the huge gains in data-efficiency. Also, the proposed 3D versions considerably outperform their 2D counterparts, which are trained on slices extracted from the 3D images.
We also measure how our methods affect the final brain tumor segmentation performance, in Table~\ref{table_brain}. All our methods outperform the baseline trained from scratch as well as their 2D counterparts, confirming the benefits of pretraining with our 3D tasks on downstream performance.
We also achieve comparable results to baselines from the BraTS challenge, and we outperform these baselines in some cases, e.g. our 3D-RPL method outperforms all baselines in terms of ET and TC dice scores. Also, our model pretrained with 3D-Exemplar, with fewer downstream training epochs, matches the result of Isensee \emph{et al.}~\cite{isensee} in terms of WT dice score, which is one of the top results on the BraTS 2018 challenge. In comparison to the supervised baseline using automatic FAST labels, we find that our results are comparable, outperforming this baseline in some cases. 
Our results in this downstream task also demonstrate the generalization ability of our 3D tasks across different domains. This is result is significant, because medical datasets are supervision-starved, e.g. images may be collected as part of clinical routine, but much fewer (high-quality) labels are produced, due to annotation costs. 

\subsection{Pancreas Tumor Segmentation Results} \label{pancreas}
\begin{figure}[ht] 
  \label{fig7} 
  \begin{minipage}{0.45\linewidth}
    \centering
    \includegraphics[width=\linewidth]{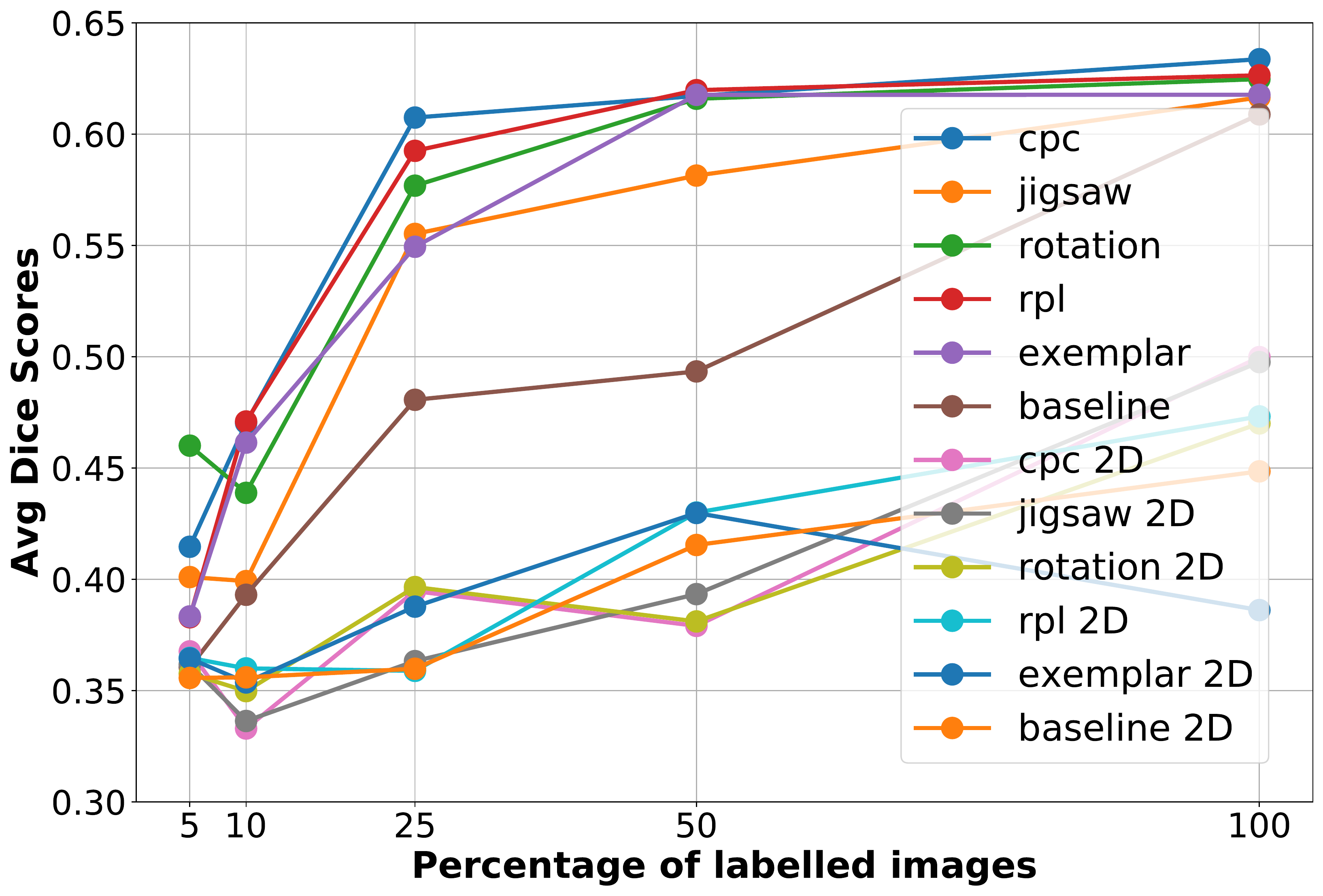}
    \caption{Data-efficient segmentation results in Pancreas. With less labeled data, the supervised baseline (brown) fails to generalize, as opposed to our methods. Also, the proposed 3D methods outperform all 2D counterparts}
    \label{plot_pancreas}
    \vspace{1ex}
  \end{minipage}\hfill
  \begin{minipage}{0.45\linewidth}
    \centering
    \includegraphics[width=\linewidth]{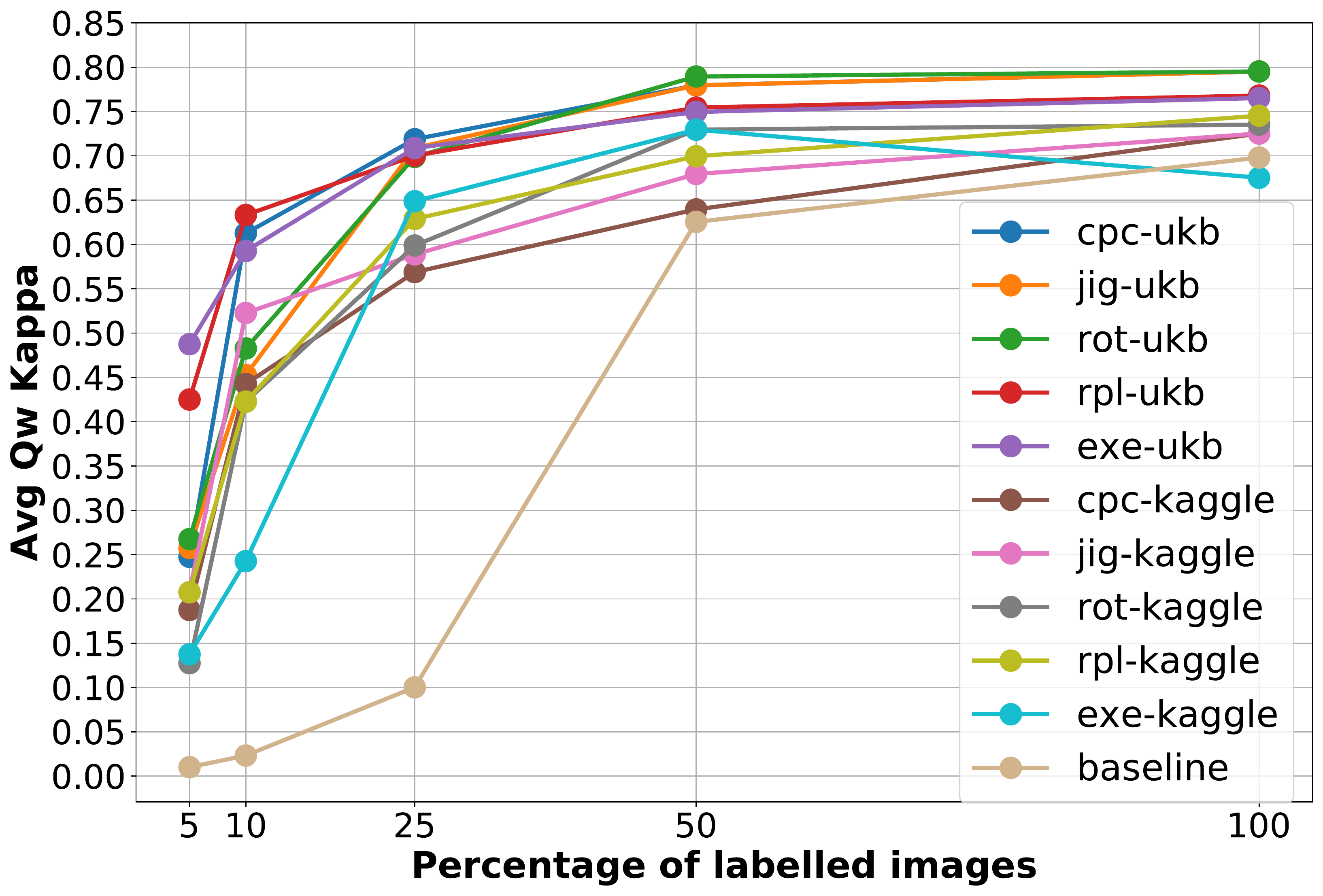}
    \caption{Data-efficient classification in Diabetic Retinopathy. With less labels, the supervised baseline (brown) fails to generalize, as opposed to pretrained models. This result is consistent with the other downstream tasks}
    \label{plot_retino}
    \vspace{1ex}
  \end{minipage} \hfill
  \begin{minipage}{0.45\linewidth}
    \centering
    \includegraphics[width=\linewidth]{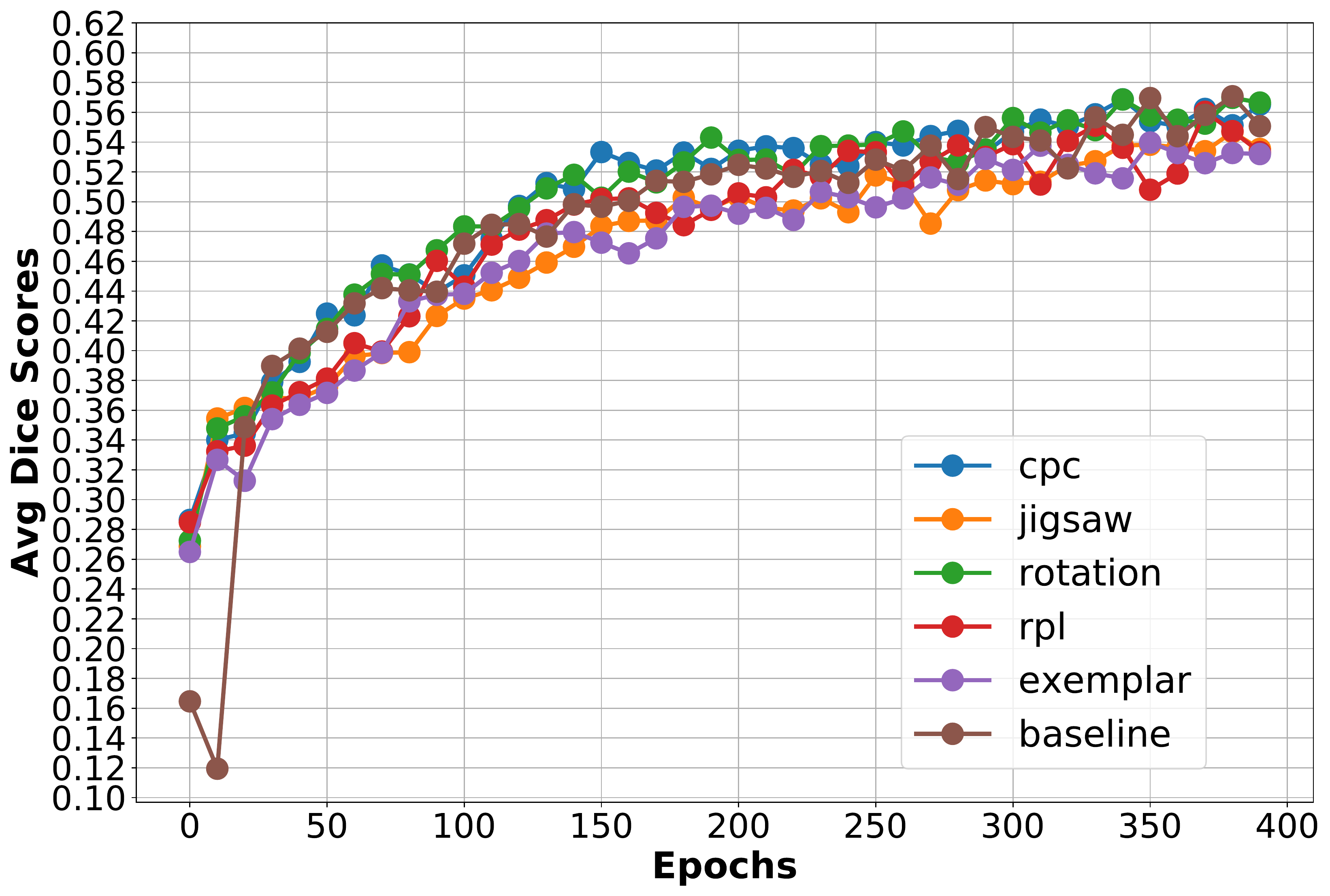}
    \caption{Speed of convergence in Pancreas segmentation. Our models converge faster than the baseline (brown)}
    \label{plot_pancreas_epochs}
  \end{minipage}\hfill
  \begin{minipage}{0.45\linewidth}
    \centering
    \includegraphics[width=\linewidth]{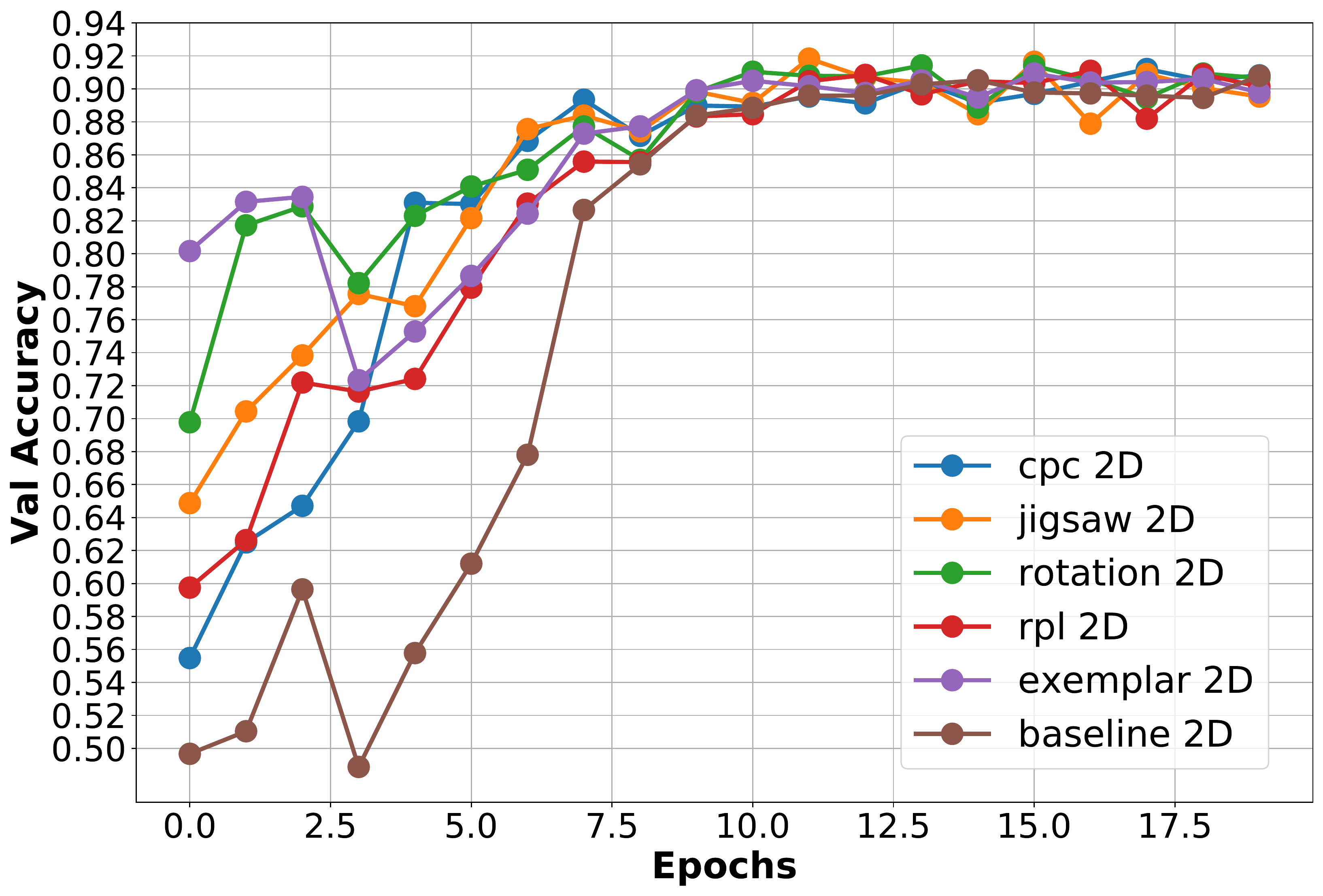}
    \caption{Speed of convergence in Retinopathy classifcation. Our models also converge faster in this task}
    \label{plot_retino_epochs}
  \end{minipage} \hfill
\end{figure}

In this downstream task, we evaluate our models on 3D CT scans of Pancreas tumor from the medical decathlon benchmarks~\cite{decathlon}. The Pancreas dataset contains annotated CT scans for $420$ cases. Each scan in this dataset contains $3$ different classes: pancreas (class 1), tumor (class 2), and background (class 0). To measure the performance on this benchmark, two dice scores are computed for classes 1 and 2. In this task, we pretrain using our proposed 3D tasks on pancreas scans \emph{without} their annotation masks. Then, we fine-tune the obtained models on subsets of annotated data to assess the gains in both data-efficiency and performance. Finally, we also compare to the baseline model trained from scratch and to 2D models, similar to the previous downstream task.
Fig.~\ref{plot_pancreas} demonstrates the gains when fine-tuning our models on 5\%, 10\%, 50\%, and 100\% of the full data size. The results obtained by our 3D methods also outperform the baselines in this task with a margin when using only few training samples, e.g. 5\% and 10\% cases. 
Another significant benefit offered by pretraining with our methods is the speed of convergence on downstream tasks. As demonstrated in Fig~\ref{plot_pancreas_epochs}, when training on the full pancreas dataset, within the first 20 epochs only, our models achieve much higher performances compared to the "from scratch" baseline. 
We should note that we evaluate this task on a held-out labeled subset of the Pancreas dataset that was not used for pretraining nor fine-tuning. We provide the full list of experimental results for this task in Appendix.

\subsection{Diabetic Retinopathy Results} \label{retino}
As part of our work, we also provide implementations for the 2D versions of the developed self-supervised methods. We showcase these implementations on the Diabetic Retinopathy 2019 Kaggle challenge~\cite{APTOS}. This dataset contains roughly $5590$ Fundus 2D images, each of which was rated by a clinician on a severity scale of $0$ to $4$. These levels define a classification task. In order to evaluate our tasks on this benchmark, we pretrain all the 2D versions of our methods using 2D Fundus images from UK Biobank~\cite{ukbio}. The retinopathy data in UK Biobank contains $170K$ images. We then fine-tune the obtained models on Kaggle data, meaning performing transfer learning. We also compare the obtained results with this transfer learning protocol to those obtained with the data-efficient evaluation protocol in~\cite{CPC2}, i.e. pretraining on the same Kaggle dataset and fine-tuning on subsets of it. 
To assess the gains in data-efficiency, we fine-tune the obtained models on subsets of labelled Kaggle data, shown in Fig.~\ref{plot_retino}. It is noteworthy that pretraining on UKB produces results that outperform those obtained when pretraining on the same Kaggle dataset. This confirms the benefits of transfer learning from a large corpus to a smaller one using our methods.
Gains in speed of convergence are also shown in Fig.~\ref{plot_retino_epochs}. In this 2D task, we achieve results consistent with the other downstream tasks, presented before. 
We should point out that we evaluate with 5-fold cross validation on this 2D dataset. The metric used in task, as in the Kaggle challenge, is the Quadratic Weighted Kappa, which measures the agreement between two ratings. Its values vary from random (0) to complete (1) agreement, and if there is less agreement than chance it may become negative. 

\section{Conclusion}
In this work, we asked whether designing 3D self-supervised tasks could benefit the learned representations from unlabeled 3D images, and found that it indeed greatly improves their downstream performance, especially when fine-tuned on only small amounts of labeled 3D data. 
We demonstrate the obtained gains by our proposed 3D algorithms in data-efficiency, performance, and speed of convergence on three different downstream tasks. Our 3D tasks outperform their 2D counterparts, hence supporting our proposal of utilizing the 3D spatial context in the design of self-supervised tasks, when operating on 3D domains. 
What is more, our results, particularly in the low-data regime, demonstrate the possibility to reduce the manual annotation effort required in the medical imaging domain, where data and annotation scarcity is an obstacle. 
Furthermore, we observe performance gains when pretraining our methods on a large unlabeled corpus, and fine-tuning them on a different smaller downstream-specific dataset. This result suggests alternatives for transfer learning from Imagenet features, which can be substantially different from the medical domain.
Finally, we open source our implementations for all 3D methods (and also their 2D versions), and we publish them to help other researchers apply our methods on other medical imaging tasks. 
This work is only a first step toward creating a set of methods that facilitate self-supervised learning research for 3D data, e.g. medical scans. We believe there is room for improvement along this line, such as designing new 3D proxy tasks, evaluating different architectural options, and including other data modalities (e.g. text) in conjunction with images/scans. 

\clearpage
%
%
\section*{Broader Impact}
Due to technological advancements in 3D data sensing, and to the growing number of its applications, the attention to machine learning algorithms that perform analysis tasks on such data has grown rapidly in the past few years. As mentioned before, 3D imaging has multitude of applications~\cite{review_DL_4_3D2}, such as in Robotics, in CAD imaging, in Geology, and in Medical Imaging. In this work, we developed multiple 3D Deep Learning algorithms, and evaluated them on multiple 3D medical imaging benchmarks. Our focus on medical imaging is motivated by the pressing demand for automatic (and instant) analysis systems, that may aid the medical community. 

Medical imaging plays an important role in patient healthcare, as it aids in disease prevention, early detection, diagnosis, and treatment. With the continuous digitization of medical images, the hope that physicians and radiologists are able to instantly analyze them with Machine Learning algorithms is slowly shaping as a reality. Achieving this has become more critical recently, as the number of patients which contracted with a novel Coronavirus, called COVID-19, reached a high record. Radiography images provide a rich and a quick diagnosis tool, because other types of tests, e.g. RT-PCR which is an RNA/DNA based test, have low sensitivity and may require hours/days of processing~\cite{COVID19_imaging}. Therefore, as imaging allows such instant insights into human body organs, it receives growing attention from both machine learning and medical communities.

Yet efforts to leverage advancements in machine learning, particularly the supervised algorithms, are often hampered by the sheer expense of expert annotation required~\cite{annotate}. Generating expert annotations of patient data at scale is non-trivial, expensive, and time-consuming, especially for 3D medical scans. Even current semi-automatic software tools fail to sufficiently address this challenge. Consequently, it is necessary to rely on annotation-efficient machine learning algorithms, such as self-supervised (unsupervised) approaches for representation learning from unlabelled data. Our work aims to provide the necessary tools for 3D image analysis, in general, and to aid physicians and radiologists in their diagnostic tasks from 3D scans, in particular. And as the main consequence of this work, the developed methods can help reduce the effort and cost of annotation required by these practitioners. In the larger goal of leveraging Machine Learning for good, our work is only a small step toward achieving this goal for patient healthcare.

\begin{ack}
This research has been supported by funding from the German Federal Ministry of Education and Research (BMBF) in the project KI-LAB-ITSE (project number 01|S19066).
This research has been conducted using the UK Biobank Resource.
\end{ack}

%
%


\small

\bibliographystyle{unsrtnat}
\bibliography{bibliography}

\begin{thebibliography}{77}
\providecommand{\natexlab}[1]{#1}
\providecommand{\url}[1]{\texttt{#1}}
\expandafter\ifx\csname urlstyle\endcsname\relax
  \providecommand{\doi}[1]{doi: #1}\else
  \providecommand{\doi}{doi: \begingroup \urlstyle{rm}\Url}\fi

\bibitem[Griffiths and Boehm(2019)]{review_DL_4_3D}
David Griffiths and Jan Boehm.
\newblock A review on deep learning techniques for 3d sensed data
  classification.
\newblock \emph{CoRR}, abs/1907.04444, 2019.
\newblock URL \url{http://arxiv.org/abs/1907.04444}.

\bibitem[Ioannidou et~al.(2017)Ioannidou, Chatzilari, Nikolopoulos, and
  Kompatsiaris]{review_DL_4_3D2}
Anastasia Ioannidou, Elisavet Chatzilari, Spiros Nikolopoulos, and Ioannis
  Kompatsiaris.
\newblock Deep learning advances in computer vision with 3d data: A survey.
\newblock \emph{ACM Computing Surveys}, 50, 06 2017.
\newblock \doi{10.1145/3042064}.

\bibitem[Su et~al.(2017 (accessed June 2, 2020))Su, Guibas, Bronstein,
  Kalogerakis, Yang, Qi, and Huang]{review_DL_4_3D3}
Hao Su, Leonidas Guibas, Michael Bronstein, Evangelos Kalogerakis, Jimei Yang,
  Charles Qi, and Qixing Huang.
\newblock \emph{3D Deep Learning}, 2017 (accessed June 2, 2020).
\newblock URL \url{http://3ddl.stanford.edu/}.

\bibitem[Gr{\"u}nberg et~al.(2017)Gr{\"u}nberg, Jimenez-del Toro, Jakab, Langs,
  Salas~Fernandez, Winterstein, Weber, and Krenn]{annotate}
Katharina Gr{\"u}nberg, Oscar Jimenez-del Toro, Andras Jakab, Georg Langs,
  Tom{\`a}s Salas~Fernandez, Marianne Winterstein, Marc-Andr{\'e} Weber, and
  Markus Krenn.
\newblock \emph{Annotating Medical Image Data}, pages 45--67.
\newblock Springer International Publishing, Cham, 2017.

\bibitem[Deng et~al.(2009)Deng, Dong, Socher, Li, Li, and
  Fei-Fei]{imagenet_cvpr09}
Jia Deng, Wei Dong, Richard Socher, Li-Jia Li, Kai Li, and Li~Fei-Fei.
\newblock Imagenet: A large-scale hierarchical image database.
\newblock In \emph{CVPR09}, Miami, FL, USA, 2009. IEEE.

\bibitem[Wang et~al.(2017)Wang, Peng, Lu, Lu, Bagheri, and
  Summers]{xray_imagenet}
Xiaosong Wang, Yifan Peng, Le~Lu, Zhiyong Lu, Mohammadhadi Bagheri, and
  Ronald~M. Summers.
\newblock Chestx-ray8: Hospital-scale chest x-ray database and benchmarks on
  weakly-supervised classification and localization of common thorax diseases.
\newblock In \emph{2017 IEEE Conference on Computer Vision and Pattern
  Recognition (CVPR)}, pages 3462--3471, 2017.

\bibitem[Rajpurkar et~al.(2017)Rajpurkar, Irvin, Zhu, Yang, Mehta, Duan, Ding,
  Bagul, Langlotz, Shpanskaya, Lungren, and Ng]{xray_imagenet2}
Pranav Rajpurkar, Jeremy Irvin, Kaylie Zhu, Brandon Yang, Hershel Mehta, Tony
  Duan, Daisy~Yi Ding, Aarti Bagul, Curtis Langlotz, Katie~S. Shpanskaya,
  Matthew~P. Lungren, and Andrew~Y. Ng.
\newblock Chexnet: Radiologist-level pneumonia detection on chest x-rays with
  deep learning.
\newblock \emph{CoRR}, abs/1711.05225, 2017.
\newblock URL \url{http://arxiv.org/abs/1711.05225}.

\bibitem[Sahlsten et~al.(2019)Sahlsten, Jaskari, Kivinen, Turunen, Jaanio,
  Hietala, and Kaski]{retinopathy_imagenet}
Jaakko Sahlsten, Joel Jaskari, Jyri Kivinen, Lauri Turunen, Esa Jaanio, Kustaa
  Hietala, and Kimmo Kaski.
\newblock Deep learning fundus image analysis for diabetic retinopathy and
  macular edema grading.
\newblock \emph{Scientific Reports}, 9, 12 2019.
\newblock \doi{10.1038/s41598-019-47181-w}.

\bibitem[Islam et~al.(2018)Islam, Hasan, and Abdullah]{retinopathy_imagenet2}
Sheikh Muhammad~Saiful Islam, Md~Mahedi Hasan, and Sohaib Abdullah.
\newblock Deep learning based early detection and grading of diabetic
  retinopathy using retinal fundus images.
\newblock \emph{CoRR}, abs/1812.10595, 2018.
\newblock URL \url{http://arxiv.org/abs/1812.10595}.

\bibitem[Torralba and {Efros}(2011)]{dataset_bias}
Antonio Torralba and Alexey~A. {Efros}.
\newblock Unbiased look at dataset bias.
\newblock In \emph{CVPR 2011}, pages 1521--1528, 2011.

\bibitem[Raghu et~al.(2019)Raghu, Zhang, Kleinberg, and Bengio]{Transfusion}
Maithra Raghu, Chiyuan Zhang, Jon Kleinberg, and Samy Bengio.
\newblock Transfusion: Understanding transfer learning for medical imaging.
\newblock In \emph{Advances in Neural Information Processing Systems 32}, pages
  3347--3357. Curran Associates, Inc., 2019.
\newblock URL
  \url{http://papers.nips.cc/paper/8596-transfusion-understanding-transfer-learning-for-medical-imaging.pdf}.

\bibitem[Eisenberg and Margulis(2011)]{medical_imaging}
Ronald Eisenberg and Alexander Margulis.
\newblock \emph{A Patient's Guide to Medical Imaging}.
\newblock New York: Oxford University Press, NY, USA, 2011.

\bibitem[Jing and Tian(2019)]{survey_self_supervised}
Longlong Jing and Yingli Tian.
\newblock Self-supervised visual feature learning with deep neural networks:
  {A} survey.
\newblock \emph{CoRR}, abs/1902.06162, 2019.
\newblock URL \url{http://arxiv.org/abs/1902.06162}.

\bibitem[Mikolov et~al.(2013)Mikolov, Chen, Corrado, and Dean]{w2v}
Tomas Mikolov, Kai Chen, Greg Corrado, and Jeffrey Dean.
\newblock Efficient estimation of word representations in vector space.
\newblock In Yoshua Bengio and Yann LeCun, editors, \emph{1st International
  Conference on Learning Representations, {ICLR} 2013, May 2-4, 2013, Workshop
  Track Proceedings}, Scottsdale, Arizona, USA, 2013. OpenReview.
\newblock URL \url{http://arxiv.org/abs/1301.3781}.

\bibitem[Doersch et~al.(2015)Doersch, Gupta, and Efros]{context_prediction}
Carl Doersch, Abhinav Gupta, and Alexei~A. Efros.
\newblock Unsupervised visual representation learning by context prediction.
\newblock In \emph{Proceedings of the 2015 IEEE International Conference on
  Computer Vision (ICCV)}, ICCV ’15, page 1422–1430, USA, 2015. IEEE
  Computer Society.
\newblock ISBN 9781467383912.
\newblock \doi{10.1109/ICCV.2015.167}.
\newblock URL \url{https://doi.org/10.1109/ICCV.2015.167}.

\bibitem[Noroozi and Favaro(2016)]{jig}
Mehdi Noroozi and Paolo Favaro.
\newblock Unsupervised learning of visual representations by solving jigsaw
  puzzles.
\newblock In Bastian Leibe, Jiri Matas, Nicu Sebe, and Max Welling, editors,
  \emph{Computer Vision -- ECCV 2016}, pages 69--84, Cham, 2016. Springer
  International Publishing.
\newblock ISBN 978-3-319-46466-4.

\bibitem[Zhang et~al.(2016)Zhang, Isola, and Efros]{color}
Richard Zhang, Phillip Isola, and Alexei~A. Efros.
\newblock Colorful image colorization.
\newblock In Bastian Leibe, Jiri Matas, Nicu Sebe, and Max Welling, editors,
  \emph{Computer Vision -- ECCV 2016}, pages 649--666, Cham, 2016. Springer
  International Publishing.
\newblock ISBN 978-3-319-46487-9.

\bibitem[Caron et~al.(2018)Caron, Bojanowski, Joulin, and Douze]{deep_cluster}
Mathilde Caron, Piotr Bojanowski, Armand Joulin, and Matthijs Douze.
\newblock Deep clustering for unsupervised learning of visual features.
\newblock In \emph{The European Conference on Computer Vision (ECCV)}, Munich,
  Germany, September 2018. Springer.

\bibitem[Gidaris et~al.(2018)Gidaris, Singh, and Komodakis]{rotations}
Spyros Gidaris, Praveer Singh, and Nikos Komodakis.
\newblock Unsupervised representation learning by predicting image rotations.
\newblock \emph{CoRR}, abs/1803.07728, 2018.
\newblock URL \url{http://arxiv.org/abs/1803.07728}.

\bibitem[Wang et~al.(2019)Wang, Zhu, Xu, and Cao]{ssl_salient_beauty}
Jiawei Wang, Shuai Zhu, Jiao Xu, and Da~Cao.
\newblock The retrieval of the beautiful: Self-supervised salient object
  detection for beauty product retrieval.
\newblock In \emph{Proceedings of the 27th ACM International Conference on
  Multimedia}, MM ’19, page 2548–2552, New York, NY, USA, 2019. Association
  for Computing Machinery.
\newblock ISBN 9781450368896.
\newblock \doi{10.1145/3343031.3356059}.
\newblock URL \url{https://doi.org/10.1145/3343031.3356059}.

\bibitem[Pathak et~al.(2016)Pathak, Kr\"ahenb\"uhl, Donahue, Darrell, and
  Efros]{context_encoders}
Deepak Pathak, Philipp Kr\"ahenb\"uhl, Jeff Donahue, Trevor Darrell, and Alexei
  Efros.
\newblock Context encoders: Feature learning by inpainting.
\newblock In \emph{The IEEE Conference on Computer Vision and Pattern
  Recognition (CVPR)}, June 2016.

\bibitem[van~den Oord et~al.(2018)van~den Oord, Li, and Vinyals]{CPC1}
A{\"{a}}ron van~den Oord, Yazhe Li, and Oriol Vinyals.
\newblock Representation learning with contrastive predictive coding.
\newblock \emph{CoRR}, abs/1807.03748, 2018.
\newblock URL \url{http://arxiv.org/abs/1807.03748}.

\bibitem[H{\'{e}}naff et~al.(2019)H{\'{e}}naff, Srinivas, Fauw, Razavi,
  Doersch, Eslami, and van~den Oord]{CPC2}
Olivier~J. H{\'{e}}naff, Aravind Srinivas, Jeffrey~De Fauw, Ali Razavi, Carl
  Doersch, S.~M.~Ali Eslami, and A{\"{a}}ron van~den Oord.
\newblock Data-efficient image recognition with contrastive predictive coding.
\newblock \emph{CoRR}, abs/1905.09272, 2019.
\newblock URL \url{http://arxiv.org/abs/1905.09272}.

\bibitem[Chen et~al.(2020)Chen, Kornblith, Norouzi, and Hinton]{chen2020simple}
Ting Chen, Simon Kornblith, Mohammad Norouzi, and Geoffrey Hinton.
\newblock A simple framework for contrastive learning of visual
  representations, 2020.

\bibitem[He et~al.(2020)He, Fan, Wu, Xie, and Girshick]{momentum_contrast}
Kaiming He, Haoqi Fan, Yuxin Wu, Saining Xie, and Ross Girshick.
\newblock Momentum contrast for unsupervised visual representation learning.
\newblock In \emph{Proceedings of the IEEE/CVF Conference on Computer Vision
  and Pattern Recognition (CVPR)}, June 2020.

\bibitem[Gutmann and Hyvärinen(2010)]{NCE}
Michael Gutmann and Aapo Hyvärinen.
\newblock Noise-contrastive estimation: A new estimation principle for
  unnormalized statistical models.
\newblock In Yee~Whye Teh and Mike Titterington, editors, \emph{Proceedings of
  the Thirteenth International Conference on Artificial Intelligence and
  Statistics}, volume~9 of \emph{Proceedings of Machine Learning Research},
  pages 297--304, Chia Laguna Resort, Sardinia, Italy, 13--15 May 2010. PMLR.
\newblock URL \url{http://proceedings.mlr.press/v9/gutmann10a.html}.

\bibitem[{Wang} and {Gupta}(2015)]{videos1}
Xiaolong {Wang} and Abhinav {Gupta}.
\newblock Unsupervised learning of visual representations using videos.
\newblock In \emph{2015 IEEE International Conference on Computer Vision
  (ICCV)}, pages 2794--2802, 2015.

\bibitem[Vondrick et~al.(2015)Vondrick, Pirsiavash, and Torralba]{videos2}
Carl Vondrick, Hamed Pirsiavash, and Antonio Torralba.
\newblock Anticipating the future by watching unlabeled video.
\newblock \emph{CoRR}, abs/1504.08023, 2015.
\newblock URL \url{http://arxiv.org/abs/1504.08023}.

\bibitem[Walker et~al.(2015)Walker, Gupta, and Hebert]{videos3}
Jacob Walker, Abhinav Gupta, and Martial Hebert.
\newblock Dense optical flow prediction from a static image.
\newblock In \emph{2015 IEEE International Conference on Computer Vision
  (ICCV)}, pages 2443--2451, 2015.

\bibitem[Purushwalkam and Gupta(2016)]{videos4}
Senthil Purushwalkam and Abhinav Gupta.
\newblock Pose from action: Unsupervised learning of pose features based on
  motion.
\newblock \emph{CoRR}, abs/1609.05420, 2016.
\newblock URL \url{http://arxiv.org/abs/1609.05420}.

\bibitem[Vondrick et~al.(2018)Vondrick, Shrivastava, Fathi, Guadarrama, and
  Murphy]{videos5}
Carl Vondrick, Abhinav Shrivastava, Alireza Fathi, Sergio Guadarrama, and Kevin
  Murphy.
\newblock Tracking emerges by colorizing videos.
\newblock In \emph{Proceedings of the European Conference on Computer Vision
  (ECCV)}, September 2018.

\bibitem[{Ji} et~al.(2013){Ji}, {Xu}, {Yang}, and {Yu}]{3DCNN_video}
S.~{Ji}, W.~{Xu}, M.~{Yang}, and K.~{Yu}.
\newblock 3d convolutional neural networks for human action recognition.
\newblock \emph{IEEE Transactions on Pattern Analysis and Machine
  Intelligence}, 35\penalty0 (1):\penalty0 221--231, 2013.

\bibitem[Dahun et~al.(2019)Dahun, Cho, and Kweon]{videos_3D_cubic_puzzles}
Kim Dahun, Donghyeon Cho, and Soo-Ok Kweon.
\newblock Self-supervised video representation learning with space-time cubic
  puzzles.
\newblock \emph{Proceedings of the AAAI Conference on Artificial Intelligence},
  33:\penalty0 8545--8552, 07 2019.
\newblock \doi{10.1609/aaai.v33i01.33018545}.

\bibitem[Jing and Tian(2018)]{videos_3D_rot}
Longlong Jing and Yingli Tian.
\newblock Self-supervised spatiotemporal feature learning by video geometric
  transformations.
\newblock \emph{CoRR}, abs/1811.11387, 2018.
\newblock URL \url{http://arxiv.org/abs/1811.11387}.

\bibitem[Liu et~al.(2018)Liu, Sinha, Unberath, Ishii, Hager, Taylor, and
  Reiter]{depth}
Xingtong Liu, Ayushi Sinha, Mathias Unberath, Masaru Ishii, Gregory~D. Hager,
  Russell~H. Taylor, and Austin Reiter.
\newblock Self-supervised learning for dense depth estimation in monocular
  endoscopy.
\newblock \emph{CoRR}, abs/1806.09521, 2018.
\newblock URL \url{http://arxiv.org/abs/1806.09521}.

\bibitem[Ye et~al.(2017)Ye, Johns, Handa, Zhang, Pratt, and Yang]{surgery}
Menglong Ye, Edward Johns, Ankur Handa, Lin Zhang, Philip Pratt, and Guang
  Yang.
\newblock Self-supervised siamese learning on stereo image pairs for depth
  estimation in robotic surgery.
\newblock In \emph{The Hamlyn Symposium on Medical Robotics}, pages 27--28, 06
  2017.
\newblock \doi{10.31256/HSMR2017.14}.

\bibitem[Li and Fan(2018)]{register}
Hongming Li and Yong Fan.
\newblock Non-rigid image registration using self-supervised fully
  convolutional networks without training data.
\newblock In \emph{2018 IEEE 15th International Symposium on Biomedical Imaging
  (ISBI 2018)}, pages 1075--1078, Washington, DC, USA, April 2018. IEEE.

\bibitem[Zhang et~al.(2017)Zhang, Wang, and Zheng]{body}
Pengyue Zhang, Fusheng Wang, and Yefeng Zheng.
\newblock Self supervised deep representation learning for fine-grained body
  part recognition.
\newblock In \emph{2017 IEEE 14th International Symposium on Biomedical Imaging
  (ISBI 2017)}, pages 578--582, Melbourne, Australia, April 2017. IEEE.

\bibitem[Jamaludin et~al.(2017)Jamaludin, Kadir, and Zisserman]{disc}
Amir Jamaludin, Timor Kadir, and Andrew Zisserman.
\newblock Self-supervised learning for spinal mris.
\newblock In \emph{Deep Learning in Medical Image Analysis and Multimodal
  Learning for Clinical Decision Support}, pages 294--302, Cham, 09 2017.
  Springer.
\newblock ISBN 978-3-319-67557-2.
\newblock \doi{10.1007/978-3-319-67558-9_34}.

\bibitem[Bai et~al.(2019)Bai, Chen, Tarroni, Duan, Guitton, Petersen, Guo,
  Matthews, and Rueckert]{cardiac_self_supervised}
Wenjia Bai, Chen Chen, Giacomo Tarroni, Jinming Duan, Florian Guitton,
  Steffen~E. Petersen, Yike Guo, Paul~M. Matthews, and Daniel Rueckert.
\newblock Self-supervised learning for cardiac mr image segmentation by
  anatomical position prediction.
\newblock In Dinggang Shen, Tianming Liu, Terry~M. Peters, Lawrence~H. Staib,
  Caroline Essert, Sean Zhou, Pew-Thian Yap, and Ali Khan, editors,
  \emph{Medical Image Computing and Computer Assisted Intervention -- MICCAI
  2019}, pages 541--549, Cham, 2019. Springer International Publishing.
\newblock ISBN 978-3-030-32245-8.

\bibitem[Yan et~al.(2019)Yan, Wang, Lu, Zhang, Harrison, Bagheri, and
  Summers]{Yan2018DeepLG}
Ke~Yan, Xiaosong Wang, Le~Lu, Ling Zhang, Adam~P. Harrison, Mohammadhadi
  Bagheri, and Ronald~M. Summers.
\newblock \emph{Deep Lesion Graph in the Wild: Relationship Learning and
  Organization of Significant Radiology Image Findings in a Diverse Large-Scale
  Lesion Database}, pages 413--435.
\newblock Springer International Publishing, Cham, 2019.
\newblock ISBN 978-3-030-13969-8.
\newblock \doi{10.1007/978-3-030-13969-8_20}.
\newblock URL \url{https://doi.org/10.1007/978-3-030-13969-8_20}.

\bibitem[Ro{\ss} et~al.(2017)Ro{\ss}, Zimmerer, Vemuri, Isensee, Bodenstedt,
  Both, Kessler, Wagner, M{\"{u}}ller, Kenngott, Speidel, Maier{-}Hein, and
  Maier{-}Hein]{endoscopic_videos}
Tobias Ro{\ss}, David Zimmerer, Anant Vemuri, Fabian Isensee, Sebastian
  Bodenstedt, Fabian Both, Philip Kessler, Martin Wagner, Beat M{\"{u}}ller,
  Hannes Kenngott, Stefanie Speidel, Klaus Maier{-}Hein, and Lena Maier{-}Hein.
\newblock Exploiting the potential of unlabeled endoscopic video data with
  self-supervised learning.
\newblock \emph{International Journal of Computer Assisted Radiology and
  Surgery}, 13, 11 2017.
\newblock \doi{10.1007/s11548-018-1772-0}.

\bibitem[Spitzer et~al.(2018)Spitzer, Kiwitz, Amunts, Harmeling, and
  Dickscheid]{Cytoarchitectonic_segmentation}
Hannah Spitzer, Kai Kiwitz, Katrin Amunts, Stefan Harmeling, and Timo
  Dickscheid.
\newblock Improving cytoarchitectonic segmentation of human brain areas with
  self-supervised siamese networks.
\newblock In Alejandro~F. Frangi, Julia~A. Schnabel, Christos Davatzikos,
  Carlos Alberola-L{\'o}pez, and Gabor Fichtinger, editors, \emph{Medical Image
  Computing and Computer Assisted Intervention -- MICCAI 2018}, pages 663--671,
  Cham, 2018. Springer International Publishing.
\newblock ISBN 978-3-030-00931-1.

\bibitem[{Tajbakhsh} et~al.(2019){Tajbakhsh}, {Hu}, {Cao}, {Yan}, {Xiao}, {Lu},
  {Liang}, {Terzopoulos}, and {Ding}]{orientation_prediction_tajbakhsh}
N.~{Tajbakhsh}, Y.~{Hu}, J.~{Cao}, X.~{Yan}, Y.~{Xiao}, Y.~{Lu}, J.~{Liang},
  D.~{Terzopoulos}, and X.~{Ding}.
\newblock Surrogate supervision for medical image analysis: Effective deep
  learning from limited quantities of labeled data.
\newblock In \emph{2019 IEEE 16th International Symposium on Biomedical Imaging
  (ISBI 2019)}, pages 1251--1255, 2019.

\bibitem[Chen et~al.(2019)Chen, Bentley, Mori, Misawa, Fujiwara, and
  Rueckert]{image_context_medical}
Liang Chen, Paul Bentley, Kensaku Mori, Kazunari Misawa, Michitaka Fujiwara,
  and Daniel Rueckert.
\newblock Self-supervised learning for medical image analysis using image
  context restoration.
\newblock \emph{Medical Image Analysis}, 58:\penalty0 101539, 2019.
\newblock ISSN 1361-8415.
\newblock \doi{https://doi.org/10.1016/j.media.2019.101539}.
\newblock URL
  \url{http://www.sciencedirect.com/science/article/pii/S1361841518304699}.

\bibitem[Jiao et~al.(2020)Jiao, Droste, Drukker, Papageorghiou, and
  Noble]{ultrasound_video}
Jianbo Jiao, Richard Droste, Lior Drukker, Aris~T. Papageorghiou, and J.~Alison
  Noble.
\newblock Self-supervised representation learning for ultrasound video.
\newblock In \emph{2020 IEEE 17th International Symposium on Biomedical Imaging
  (ISBI)}, pages 1847--1850, 2020.

\bibitem[Taleb et~al.(2019)Taleb, Lippert, Klein, and Nabi]{multimodal_puzzles}
Aiham Taleb, Christoph Lippert, Tassilo Klein, and Moin Nabi.
\newblock Multimodal self-supervised learning for medical image analysis, 2019.

\bibitem[Blendowski et~al.(2019)Blendowski, Nickisch, and
  Heinrich]{image_context_medical2}
Maximilian Blendowski, Hannes Nickisch, and Mattias~P. Heinrich.
\newblock How to learn from unlabeled volume data: Self-supervised 3d context
  feature learning.
\newblock In Dinggang Shen, Tianming Liu, Terry~M. Peters, Lawrence~H. Staib,
  Caroline Essert, Sean Zhou, Pew-Thian Yap, and Ali Khan, editors,
  \emph{Medical Image Computing and Computer Assisted Intervention -- MICCAI
  2019}, pages 649--657, Cham, 2019. Springer International Publishing.
\newblock ISBN 978-3-030-32226-7.

\bibitem[Chaitanya et~al.(2020)Chaitanya, Erdil, Karani, and
  Konukoglu]{chaitanya2020contrastive}
Krishna Chaitanya, Ertunc Erdil, Neerav Karani, and Ender Konukoglu.
\newblock Contrastive learning of global and local features for medical image
  segmentation with limited annotations, 2020.

\bibitem[Zhou et~al.(2019)Zhou, Sodha, Rahman~Siddiquee, Feng, Tajbakhsh,
  Gotway, and Liang]{ssl_models_genesis}
Zongwei Zhou, Vatsal Sodha, Md~Mahfuzur Rahman~Siddiquee, Ruibin Feng, Nima
  Tajbakhsh, Michael~B. Gotway, and Jianming Liang.
\newblock Models genesis: Generic autodidactic models for 3d medical image
  analysis.
\newblock In \emph{Medical Image Computing and Computer Assisted Intervention
  -- MICCAI 2019}, pages 384--393, Cham, 2019. Springer International
  Publishing.
\newblock ISBN 978-3-030-32251-9.

\bibitem[Zhuang et~al.(2019)Zhuang, Li, Hu, Ma, Yang, and Zheng]{rubik}
Xinrui Zhuang, Yuexiang Li, Yifan Hu, Kai Ma, Yujiu Yang, and Yefeng Zheng.
\newblock Self-supervised feature learning for 3d medical images by playing a
  rubik's cube.
\newblock In Dinggang Shen, Tianming Liu, Terry~M. Peters, Lawrence~H. Staib,
  Caroline Essert, Sean Zhou, Pew-Thian Yap, and Ali Khan, editors,
  \emph{Medical Image Computing and Computer Assisted Intervention -- MICCAI
  2019}, pages 420--428, Cham, 2019. Springer International Publishing.
\newblock ISBN 978-3-030-32251-9.

\bibitem[Zhu et~al.(2020)Zhu, Li, Hu, Ma, Zhou, and Zheng]{rubik2}
Jiuwen Zhu, Yuexiang Li, Yifan Hu, Kai Ma, S.~Kevin Zhou, and Yefeng Zheng.
\newblock Rubik’s cube+: A self-supervised feature learning framework for 3d
  medical image analysis.
\newblock \emph{Medical Image Analysis}, 64:\penalty0 101746, 2020.
\newblock ISSN 1361-8415.
\newblock \doi{https://doi.org/10.1016/j.media.2020.101746}.
\newblock URL
  \url{http://www.sciencedirect.com/science/article/pii/S1361841520301109}.

\bibitem[Cho et~al.(2014)Cho, van Merri{\"e}nboer, G{\"{u}}l{\c{c}}ehre,
  Bahdanau, Bougares, Schwenk, and Bengio]{GRU}
Kyunghyun Cho, Bart van Merri{\"e}nboer, {\c{C}}aglar G{\"{u}}l{\c{c}}ehre,
  Dzmitry Bahdanau, Fethi Bougares, Holger Schwenk, and Yoshua Bengio.
\newblock Learning phrase representations using {RNN} encoder{--}decoder for
  statistical machine translation.
\newblock In \emph{Proceedings of the 2014 Conference on Empirical Methods in
  Natural Language Processing ({EMNLP})}, pages 1724--1734, Doha, Qatar,
  October 2014. Association for Computational Linguistics.
\newblock \doi{10.3115/v1/D14-1179}.
\newblock URL \url{https://www.aclweb.org/anthology/D14-1179}.

\bibitem[van~den Oord et~al.(2016)van~den Oord, Kalchbrenner, Espeholt,
  kavukcuoglu, Vinyals, and Graves]{pixel_cnn}
Aaron van~den Oord, Nal Kalchbrenner, Lasse Espeholt, koray kavukcuoglu, Oriol
  Vinyals, and Alex Graves.
\newblock Conditional image generation with pixelcnn decoders.
\newblock In D.~D. Lee, M.~Sugiyama, U.~V. Luxburg, I.~Guyon, and R.~Garnett,
  editors, \emph{Advances in Neural Information Processing Systems 29}, pages
  4790--4798. Curran Associates, Inc., 2016.
\newblock URL
  \url{http://papers.nips.cc/paper/6527-conditional-image-generation-with-pixelcnn-decoders.pdf}.

\bibitem[Stollenga et~al.(2015)Stollenga, Byeon, Liwicki, and
  Schmidhuber]{pyramid_LSTM}
Marijn~F. Stollenga, Wonmin Byeon, Marcus Liwicki, and J{\"{u}}rgen
  Schmidhuber.
\newblock Parallel multi-dimensional lstm, with application to fast biomedical
  volumetric image segmentation.
\newblock \emph{CoRR}, abs/1506.07452, 2015.
\newblock URL \url{http://arxiv.org/abs/1506.07452}.

\bibitem[Van Den~Oord et~al.(2016)Van Den~Oord, Kalchbrenner, and
  Kavukcuoglu]{pixel_rnn}
A\"{a}ron Van Den~Oord, Nal Kalchbrenner, and Koray Kavukcuoglu.
\newblock Pixel recurrent neural networks.
\newblock In \emph{Proceedings of the 33rd International Conference on
  International Conference on Machine Learning - Volume 48}, ICML’16, page
  1747–1756. JMLR.org, 2016.

\bibitem[Dosovitskiy et~al.(2014)Dosovitskiy, Springenberg, Riedmiller, and
  Brox]{exemplar}
Alexey Dosovitskiy, Jost~T. Springenberg, Martin Riedmiller, and Thomas Brox.
\newblock Discriminative unsupervised feature learning with convolutional
  neural networks.
\newblock In \emph{Advances in Neural Information Processing Systems 27
  (NIPS)}, 2014.
\newblock URL
  \url{http://lmb.informatik.uni-freiburg.de/Publications/2014/DB14b}.

\bibitem[Wang and Gupta(2015)]{exemplar_triplet}
Xiaolong Wang and Abhinav Gupta.
\newblock Unsupervised learning of visual representations using videos.
\newblock In \emph{2015 IEEE International Conference on Computer Vision
  (ICCV)}, pages 2794--2802, 2015.

\bibitem[Schroff et~al.(2015)Schroff, Kalenichenko, and
  Philbin]{FaceNet_triplet}
Florian Schroff, Dmitry Kalenichenko, and James Philbin.
\newblock Facenet: A unified embedding for face recognition and clustering.
\newblock In \emph{2015 IEEE Conference on Computer Vision and Pattern
  Recognition (CVPR)}, pages 815--823, 2015.

\bibitem[Goyal et~al.(2019)Goyal, Mahajan, Mulam, and
  Misra]{scaling_self_supervised}
Priya Goyal, Dhruv Mahajan, Harikrishna Mulam, and Ishan Misra.
\newblock Scaling and benchmarking self-supervised visual representation
  learning.
\newblock In \emph{The IEEE International Conference on Computer Vision
  (ICCV)}, pages 6390--6399, October 2019.
\newblock \doi{10.1109/ICCV.2019.00649}.

\bibitem[Menze et~al.(2015)Menze, Jakab, Bauer, Kalpathy-Cramer, Farahani,
  Kirby, Burren, and et~al.]{brats1}
Bjoern~H. Menze, Andras Jakab, Stefan Bauer, Jayashree Kalpathy-Cramer, Keyvan
  Farahani, Justin Kirby, Yuliya Burren, and et~al.
\newblock The multimodal brain tumor image segmentation benchmark (brats).
\newblock \emph{IEEE Transactions on Medical Imaging}, 34\penalty0
  (10):\penalty0 1993--2024, 2015.

\bibitem[Bakas et~al.(2017)Bakas, Akbari, Sotiras, Bilello, Rozycki, Kirby,
  Freymann, Farahani, and Davatzikos]{brats2}
Spyridon Bakas, Hamed Akbari, Aristeidis Sotiras, Michel Bilello, Martin
  Rozycki, Justin~S. Kirby, John~B. Freymann, Keyvan Farahani, and Christos
  Davatzikos.
\newblock Advancing the cancer genome atlas glioma mri collections with expert
  segmentation labels and radiomic features.
\newblock \emph{Scientific Data}, 4:\penalty0 170117 EP --, 09 2017.

\bibitem[Sudlow et~al.(2015)Sudlow, Gallacher, Allen, Beral, Burton, Danesh,
  Downey, Elliott, Green, Landray, Liu, Matthews, Ong, Pell, Silman, Young,
  Sprosen, Peakman, and Collins]{ukbio}
Cathie Sudlow, John Gallacher, Naomi Allen, Valerie Beral, Paul Burton, John
  Danesh, Paul Downey, Paul Elliott, Jane Green, Martin Landray, Bette Liu,
  Paul Matthews, Giok Ong, Jill Pell, Alan Silman, Alan Young, Tim Sprosen, Tim
  Peakman, and Rory Collins.
\newblock Uk biobank: An open access resource for identifying the causes of a
  wide range of complex diseases of middle and old age.
\newblock \emph{PLOS Medicine}, 12\penalty0 (3):\penalty0 1--10, 03 2015.
\newblock \doi{10.1371/journal.pmed.1001779}.
\newblock URL \url{https://doi.org/10.1371/journal.pmed.1001779}.

\bibitem[Woolrich et~al.(2009)Woolrich, Jbabdi, Patenaude, Chappell, Makni,
  Behrens, Beckmann, Jenkinson, and Smith]{fsl_fast}
Mark~W. Woolrich, Saad Jbabdi, Brian Patenaude, Michael Chappell, Salima Makni,
  Timothy Behrens, Christian Beckmann, Mark Jenkinson, and Stephen~M. Smith.
\newblock Bayesian analysis of neuroimaging data in fsl.
\newblock \emph{NeuroImage}, 45\penalty0 (1, Supplement 1):\penalty0 S173 --
  S186, 2009.
\newblock ISSN 1053-8119.
\newblock \doi{https://doi.org/10.1016/j.neuroimage.2008.10.055}.
\newblock URL
  \url{http://www.sciencedirect.com/science/article/pii/S1053811908012044}.
\newblock Mathematics in Brain Imaging.

\bibitem[Isensee et~al.(2018)Isensee, Kickingereder, Wick, Bendszus, and
  Maier-Hein]{isensee}
Fabian Isensee, Philipp Kickingereder, Wolfgang Wick, Martin Bendszus, and
  Klaus~H Maier-Hein.
\newblock No new-net.
\newblock In \emph{International MICCAI Brainlesion Workshop}, pages 234--244,
  Granada, Spain, 2018. Springer.

\bibitem[Popli et~al.(2018)Popli, Agarwal, and Pillai]{Popli}
Anmol Popli, Manu Agarwal, and G.N. Pillai.
\newblock Automatic brain tumor segmentation using u-net based 3d fully
  convolutional network.
\newblock In \emph{Pre-Conference Proceedings of the 7th MICCAI BraTS
  Challenge}, pages 374--382. Springer, 2018.

\bibitem[Baid et~al.(2018)Baid, Mahajan, Talbar, Rane, Thakur, Moiyadi, Thakur,
  and Gupta]{Baid}
Ujjwal Baid, Abhishek Mahajan, Sanjay Talbar, Swapnil Rane, Siddhesh Thakur,
  Aliasgar Moiyadi, Meenakshi Thakur, and Sudeep Gupta.
\newblock Gbm segmentation with 3d u-net and survival prediction with
  radiomics.
\newblock In \emph{International MICCAI Brainlesion Workshop}, pages 28--35.
  Springer, 2018.

\bibitem[Chandra et~al.(2018)Chandra, Vakalopoulou, Fidon, Battistella,
  Estienne, Sun, Robert, Deutch, and Paragios]{Chandra}
Siddhartha Chandra, Maria Vakalopoulou, Lucas Fidon, Enzo Battistella, Theo
  Estienne, Roger Sun, Charlotte Robert, Eric Deutch, and Nikos Paragios.
\newblock Context aware 3-d residual networks for brain tumor segmentation.
\newblock In \emph{International MICCAI Brainlesion Workshop}, pages 74--82.
  Springer, 2018.

\bibitem[Ronneberger et~al.(2015)Ronneberger, Fischer, and Brox]{UNET}
Olaf Ronneberger, Philipp Fischer, and Thomas Brox.
\newblock U-net: Convolutional networks for biomedical image segmentation.
\newblock In Nassir Navab, Joachim Hornegger, William~M. Wells, and
  Alejandro~F. Frangi, editors, \emph{Medical Image Computing and
  Computer-Assisted Intervention -- MICCAI 2015}, pages 234--241, Cham, 2015.
  Springer International Publishing.
\newblock ISBN 978-3-319-24574-4.

\bibitem[Simpson et~al.(2019)Simpson, Antonelli, Bakas, Bilello, Farahani, van
  Ginneken, Kopp{-}Schneider, Landman, Litjens, Menze, Ronneberger, Summers,
  Bilic, Christ, Do, Gollub, Golia{-}Pernicka, Heckers, Jarnagin, McHugo,
  Napel, Vorontsov, Maier{-}Hein, and Cardoso]{decathlon}
Amber~L. Simpson, Michela Antonelli, Spyridon Bakas, Michel Bilello, Keyvan
  Farahani, Bram van Ginneken, Annette Kopp{-}Schneider, Bennett~A. Landman,
  Geert J.~S. Litjens, Bjoern~H. Menze, Olaf Ronneberger, Ronald~M. Summers,
  Patrick Bilic, Patrick~Ferdinand Christ, Richard K.~G. Do, Marc Gollub,
  Jennifer Golia{-}Pernicka, Stephan Heckers, William~R. Jarnagin, Maureen
  McHugo, Sandy Napel, Eugene Vorontsov, Lena Maier{-}Hein, and M.~Jorge
  Cardoso.
\newblock A large annotated medical image dataset for the development and
  evaluation of segmentation algorithms.
\newblock \emph{CoRR}, abs/1902.09063, 2019.
\newblock URL \url{http://arxiv.org/abs/1902.09063}.

\bibitem[APT(2019)]{APTOS}
Aptos 2019.
\newblock \url{https://www.kaggle.com/c/aptos2019-blindness-detection/}, 2019.
\newblock Accessed: 2020-11-02.

\bibitem[Manna et~al.(2020)Manna, Wruble, Maron, Toussie, Voutsinas,
  Finkelstein, Cedillo, Diamond, Eber, Jacobi, Chung, and
  Bernheim]{COVID19_imaging}
Sayan Manna, Jill Wruble, Samuel~Z Maron, Danielle Toussie, Nicholas Voutsinas,
  Mark Finkelstein, Mario~A Cedillo, Jamie Diamond, Corey Eber, Adam Jacobi,
  Michael Chung, and Adam Bernheim.
\newblock Covid-19: A multimodality review of radiologic techniques, clinical
  utility, and imaging features.
\newblock \emph{Radiology: Cardiothoracic Imaging}, 2\penalty0 (3):\penalty0
  e200210, 2020.
\newblock \doi{10.1148/ryct.2020200210}.
\newblock URL \url{https://doi.org/10.1148/ryct.2020200210}.

\bibitem[tensorflow.org(2020 (accessed June 3, 2020))]{tf}
tensorflow.org.
\newblock \emph{Tensorflow v2.1}, 2020 (accessed June 3, 2020).
\newblock URL
  \url{https://www.tensorflow.org/versions/r2.1/api_docs/python/tf}.

\bibitem[Huang et~al.(2017)Huang, Liu, Maaten, and Weinberger]{densenet}
Gao Huang, Zhuang Liu, Laurens Van~Der Maaten, and Kilian~Q. Weinberger.
\newblock Densely connected convolutional networks.
\newblock In \emph{2017 IEEE Conference on Computer Vision and Pattern
  Recognition (CVPR)}, pages 2261--2269, 2017.

\bibitem[Kingma and Ba(2014)]{Adam_supp}
Diederik~P. Kingma and Jimmy Ba.
\newblock Adam: A method for stochastic optimization, 2014.
\newblock URL \url{http://arxiv.org/abs/1412.6980}.
\newblock cite arxiv:1412.6980Comment: Published as a conference paper at the
  3rd International Conference for Learning Representations, San Diego, 2015.

\bibitem[Kolesnikov et~al.(2019)Kolesnikov, Zhai, and Beyer]{revisiting}
Alexander Kolesnikov, Xiaohua Zhai, and Lucas Beyer.
\newblock Revisiting self-supervised visual representation learning.
\newblock In \emph{The IEEE Conference on Computer Vision and Pattern
  Recognition (CVPR)}, June 2019.

\bibitem[Snoek et~al.(2005)Snoek, Worring, and Smeulders]{early_fusion}
Cees G.~M. Snoek, Marcel Worring, and Arnold W.~M. Smeulders.
\newblock Early versus late fusion in semantic video analysis.
\newblock In \emph{Proceedings of the 13th Annual ACM International Conference
  on Multimedia}, MULTIMEDIA '05, pages 399--402, New York, NY, USA, 2005. ACM.
\newblock ISBN 1-59593-044-2.
\newblock \doi{10.1145/1101149.1101236}.
\newblock URL \url{http://doi.acm.org/10.1145/1101149.1101236}.

\end{thebibliography}

\clearpage

\begin{appendices}

\section{Implementation and training details for all tasks} \label{appendix1}
It is noteworthy that our attached implementations are flexible enough to allow for evaluating several types of network architectures for encoders, decoders, and classifiers. We also provide implementations for multiple losses, augmentation techniques, and evaluation metrics. More information can be found in the \texttt{README.md} file in our attached code-base. We rely on \texttt{tensorflow v2.1}~\cite{tf} with \texttt{Keras API} in our implementations. Below, we provide the training details we used in implementing our 3D self-supervised tasks (and their 2D counterparts), and when fine-tuning them in subsequent downstream tasks.

\paragraph{Architecture details.} 
For all 3D encoders $g_{enc}$, which are pretrained with our 3D self-supervised tasks and later fine-tuned on 3D segmentation tasks, we use a 3D U-Net~\cite{UNET}-based encoder (the downward path), which consists of five levels of residual convolutional blocks. The numbers of filters in these blocks are ${32,64,128,256,512}$, respectively. The U-Net decoder (the upward path) is added in the downstream fine-tuning stage, and it includes five levels of deconvolutional blocks with skip connections from the U-Net encoder blocks.
For the 2D encoders, we use a standard Densenet-121~\cite{densenet} architecture, which is fine-tuned later on 2D classification tasks. When training our 3D self-supervised tasks, we follow~\cite{chen2020simple} in adding nonlinear transformations (a hidden layer with ReLU activation) before the final classification layers. These classification layers are removed when fine-tuning the resulting encoders $g_{enc}$ in downstream tasks.

\paragraph{Optimization details.} 
In all self-supervised and downstream tasks, we use Adam~\cite{Adam_supp} optimizer to train the models. The initial learning rate we use is $0.001$ in 3D self-supervised tasks, $0.00001$ in 3D segmentation tasks, $0.0005$ in 2D self-supervised tasks, and $0.00005$ in 2D classification tasks. 
When we fine-tune our pretrained encoders in subsequent downstream tasks, we follow a warm-up procedure inspired from~\cite{revisiting} by keeping the encoder weights frozen for a number of initial warm-up epochs while the network decoders or classifiers are trained. These warm-up epochs are 5 in 2D classification tasks, and 25 epochs in 3D segmentation tasks. The alternative options we evaluated were: 1) fine-tuning the encoder directly with a randomly initialized decoder, 2) keeping the encoder frozen throughout the training procedure. And the \nth{3} option we followed in the end was the hybrid approach of warm-up epochs described above, as it provided a performance boost over the other alternatives. For segmentation tasks, in particular, where a decoder is used in the architecture, these warm-up epochs prove indispensable. Otherwise, training the whole model with a randomly initialized decoder, while the encoder is not frozen, may harm the encoder representations.

\paragraph{Input preprocessing.} 
For all input scans, we perform the following preprocessing steps:
\begin{itemize}
    \item In self-supervised pretraining using 3D scans, we find the boundaries of the brain or the pancreas along each axis, and then we crop the remaining empty parts from the scan. This step reduces the amount of empty background voxels, as they might confuse patch-based self-supervised methods with no additional semantic information. This step is not performed when fine-tuning on 3D downstream tasks.
    \item Then, we resize each 3D image from BraTS or Pancreas to a unified resolution of $128\times128\times128$, and to the resolution $224 \times 224$ for 2D images from Diabetic Retinopathy.
    \item Then, each image's intensity values are normalized by scaling them to the range $[0,1]$.
\end{itemize}

\paragraph{Processing multimodal inputs.}
In the first downstream task of brain tumor segmentation with 3D multimodal MRI, we pretrain using the UK Biobank~\cite{ukbio} corpus, as mentioned earlier. Brain scans obtained from UKB contain 2 MRI modalities (T1 and T2-Flair), which are co-registered. This allows us to stack these 2 modalities as color channels in each input sample, similar to RGB channels. This form of early fusion~\cite{early_fusion} of MRI modalities is common when they are registered, and is a practical solution for combining all information that exist in these modalities. However, as mentioned earlier, we use the BraTS~\cite{brats1,brats2} dataset for fine-tuning, and each scan consists of 4 different MRI modalities, as opposed to only 2 in UKB that is used for pretraining. This difference only affects the input layer of the pretrained encoder, as fine-tuning on an incompatible number of input channels causes this process of fine-tuning to fail. We resolve this issue by duplicating (copying) the weights of \emph{only} the pretrained input layer. This minor modification only adds a few additional parameters to the input layer, but allows us to leverage its weights. The other alternative for this solution would have been to discard the weights of this input layer, and initialize the rest of the model layers from pretrained models normally. But we believe our solution for this issue takes advantage of any useful information encoded in these weights. 
This multimodal inputs problem does not occur in the other downstream tasks, as the inputs include only one modality/channel.

\paragraph{Task specific training details.}
\begin{itemize}
    \item \textbf{3D-CPC and 3D-Exe}: we use latent representation code size of 1024 in these tasks.
    \item \textbf{3D-Jig and 3D-RPL}: We split the input 3D images into $3 \times 3 \times 3$ patches in this task. We apply a random jitter of 3 pixels per side (axis). 
    \item \textbf{Patch-based tasks (3D-CPC, 3D-RPL, 3D-Jig)}: each extracted patch is represented using an embedding vector of size 64.
    \item \textbf{3D-Exe}: the $\alpha$ value used for the triplet loss is $1.0$.
    \item \textbf{3D-Jig}: the complexity of the Jigsaw puzzle solving task relies on the number of permutations used in generating the puzzles, i.e. the more permutations used, the harder the task is to solve. We follow the Hamming distance-based algorithm from~\cite{jig} in sampling the permutations for this task. However, in our 3D puzzles task, we sample permutations that are more complex with 27 different entries. This algorithm works as follows: we sample a subset of 1000 permutations which are selected based on their Hamming distance, i.e., the number of different tile locations between 2 permutations. When generating permutations, we ensure that the average Hamming distance across permutations is kept as high as possible. This results in a set of permutations (classes) that are as far as possible from each other.
\end{itemize}

\paragraph{Augmentation in Exemplar.} As mentioned earlier, we apply the following 3D transformations in Exemplar: random flipping along an arbitrary axis, random rotation along an arbitrary axis, random brightness and contrast, and random zooming. These augmentations are utilized to produce the positive samples. We vary the percentages of applying these augmentations using these factors: $\alpha=0.5$ for random rotations, $\beta=0.5$ for color distortions (brightness and contrast), and $\gamma=0.2$ for random zooming. When trying to omit a certain augmentation from the list above, we observe a drop in downstream performance. This is consistent with the findings of~\cite{chen2020simple}.
However, performing such transformations for high percentages is time-consuming, hence the reduced rates to $50\%$. 
Conducting a more thorough analysis of what \emph{types} of augmentations are desirable is a future work. 

\clearpage
\section{Detailed experimental results}
\begin{figure}[ht]
     \centering
     \begin{subfigure}[b]{0.32\textwidth}
         \centering
         \includegraphics[width=\linewidth]{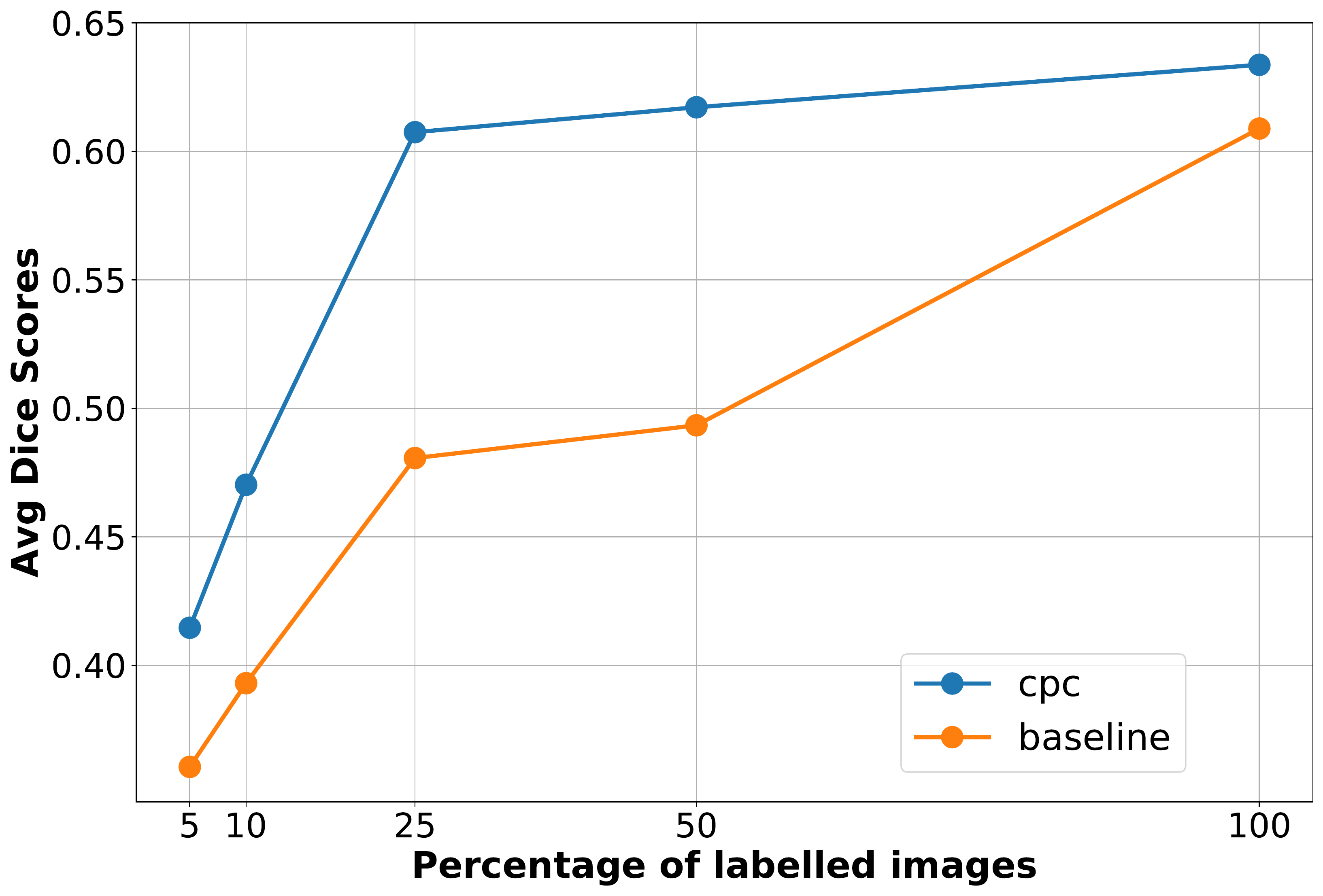}
         \caption{CPC 3D vs. baseline}
     \end{subfigure}
     \hfill
     \begin{subfigure}[b]{0.32\textwidth}
         \centering
         \includegraphics[width=\linewidth]{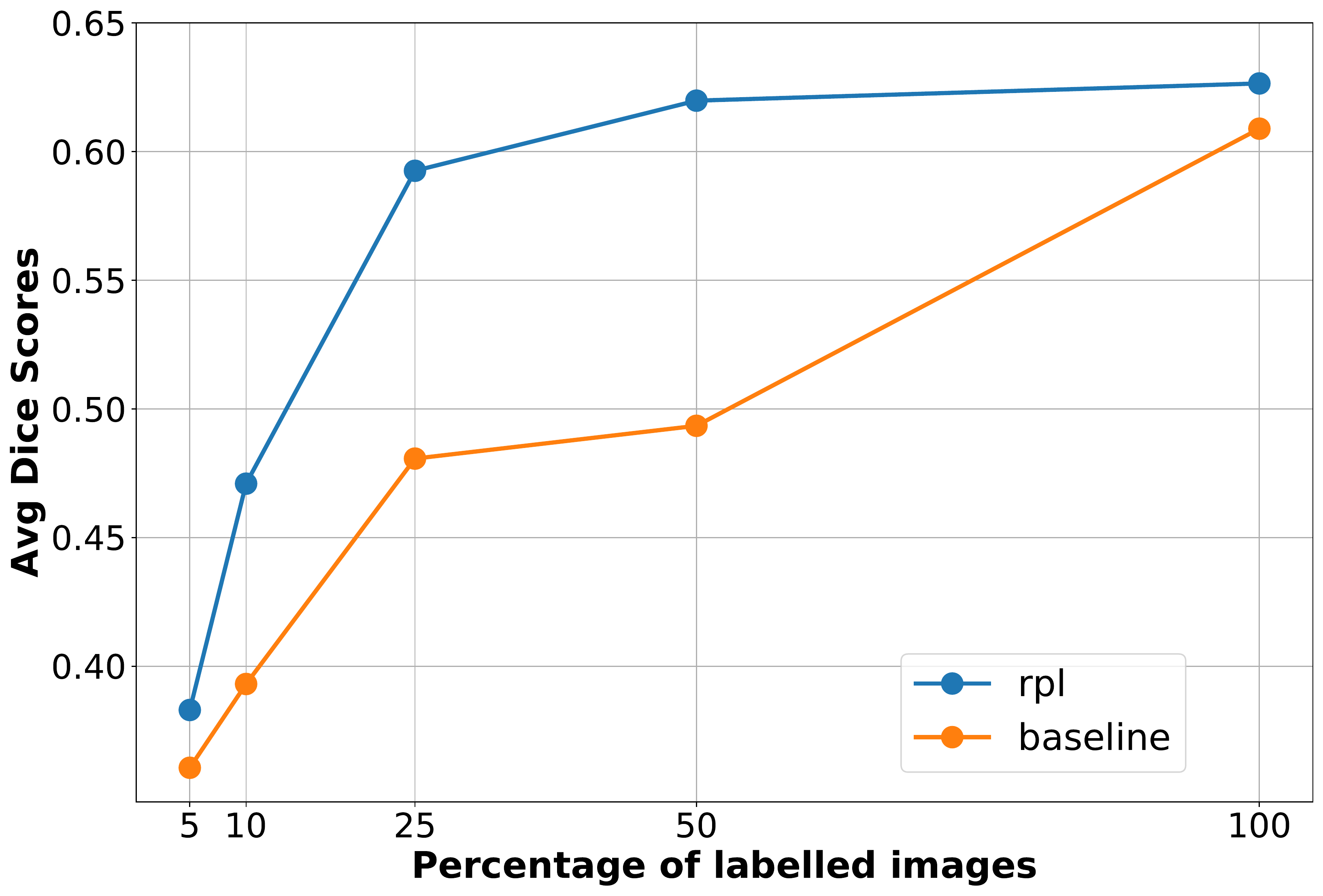}
         \caption{RPL 3D vs. baseline}
     \end{subfigure}
     \hfill
     \begin{subfigure}[b]{0.32\textwidth}
         \centering
         \includegraphics[width=\linewidth]{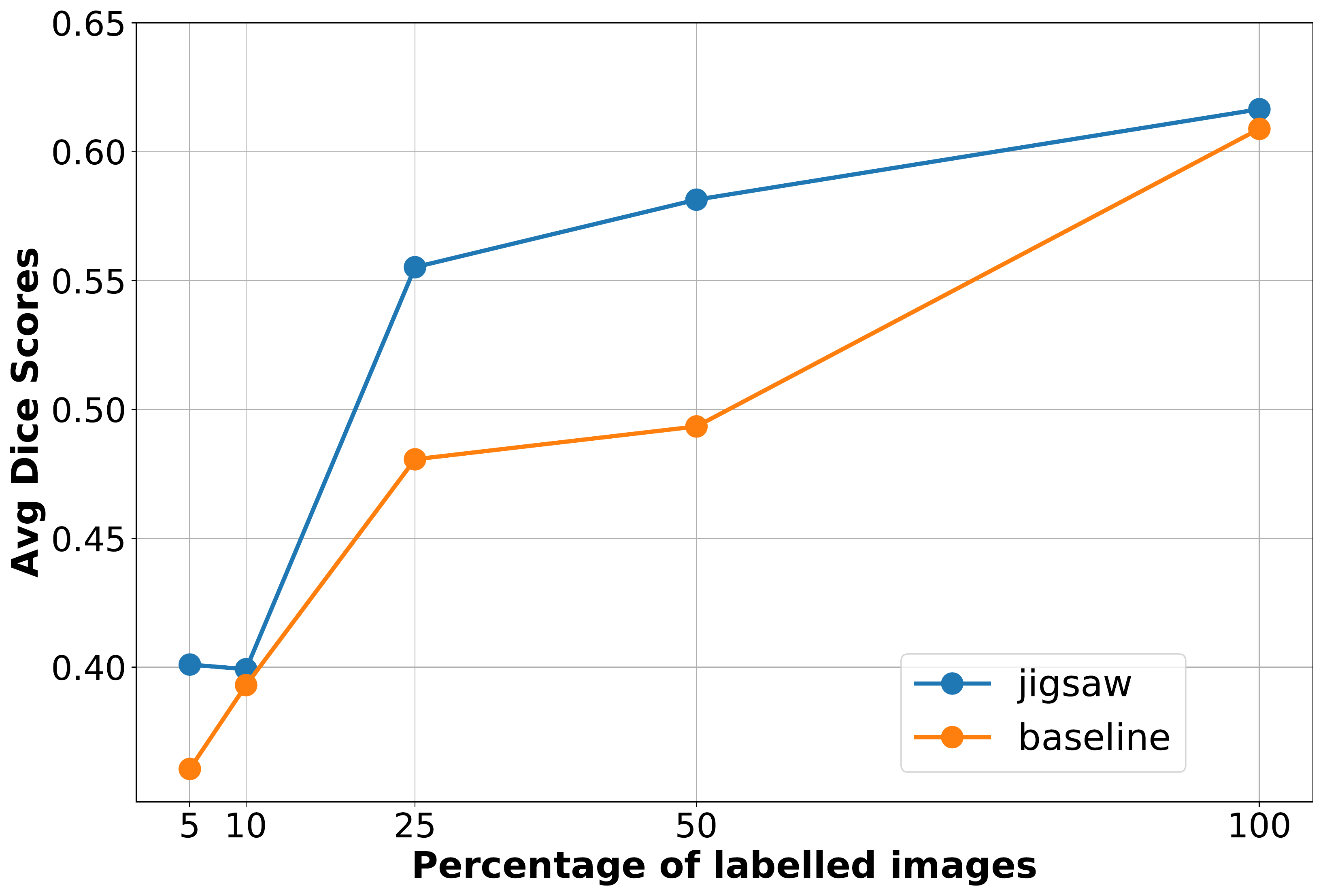}
         \caption{Jigsaw 3D vs. baseline}
     \end{subfigure}
     \vskip\baselineskip
     \begin{subfigure}[b]{0.32\textwidth}
         \centering
         \includegraphics[width=\linewidth]{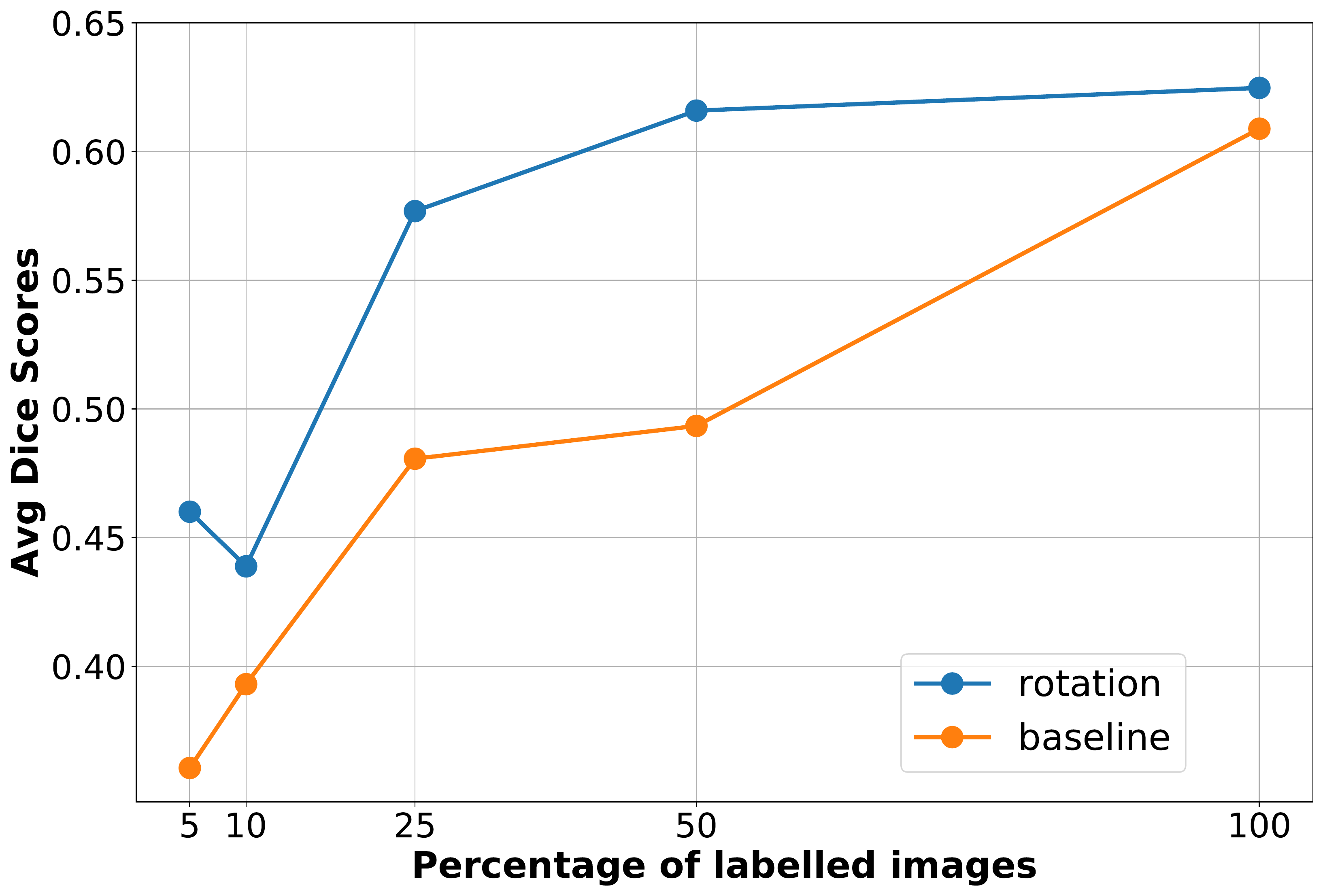}
         \caption{Rotation 3D vs. baseline}
     \end{subfigure}
     \hfill
     \begin{subfigure}[b]{0.32\textwidth}
         \centering
         \includegraphics[width=\linewidth]{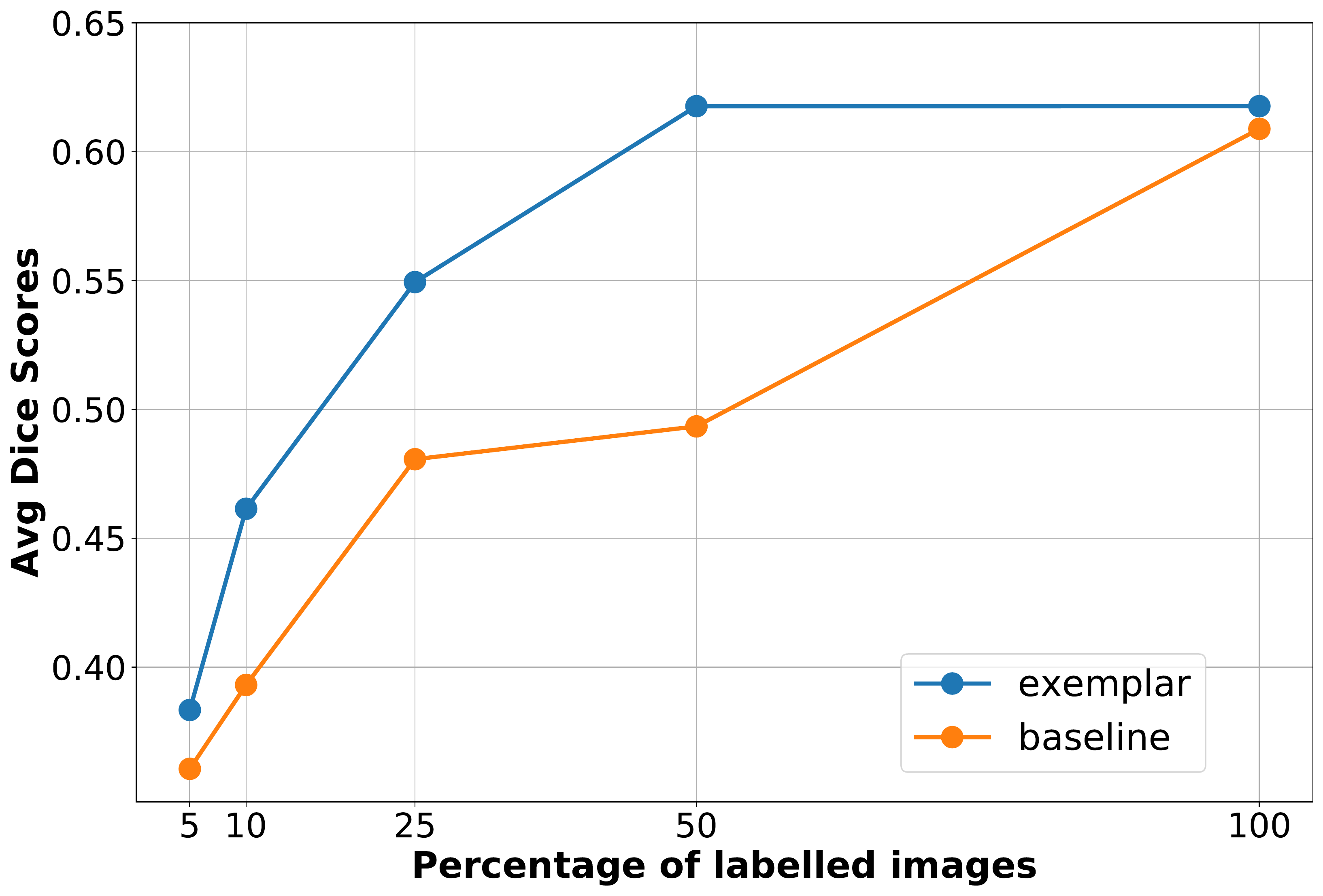}
         \caption{Exemplar 3D vs. baseline}
     \end{subfigure}
    \caption{Pancreas segmentation: Detailed data-efficiency results per method (blue) vs. the supervised baseline (orange). Our methods consistently outperform the baseline in low-data cases}
\end{figure}

\begin{figure}[htb]
     \centering
     \begin{subfigure}[b]{0.32\textwidth}
         \centering
         \includegraphics[width=\linewidth]{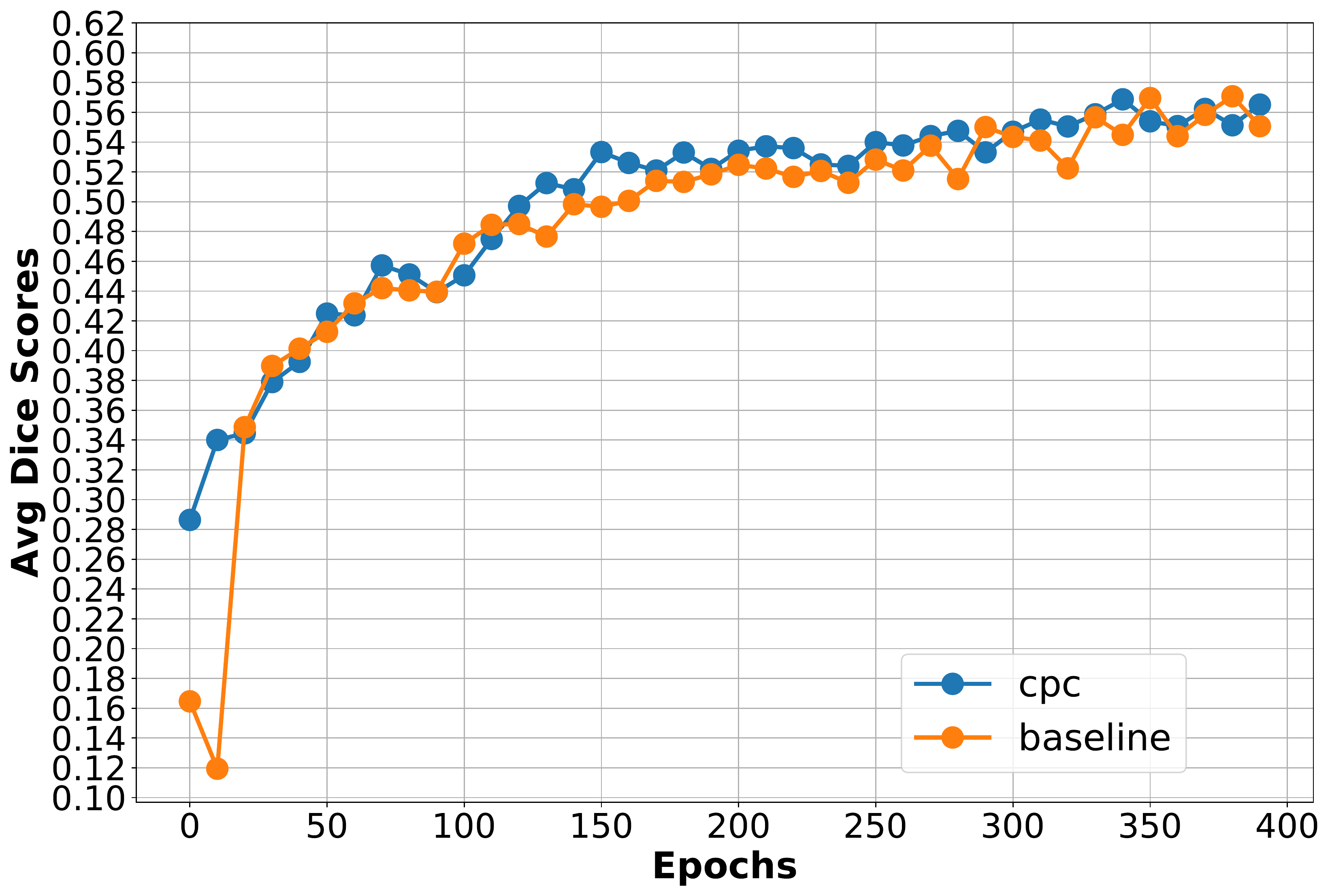}
         \caption{CPC 3D vs. baseline}
     \end{subfigure}
     \hfill
     \begin{subfigure}[b]{0.32\textwidth}
         \centering
         \includegraphics[width=\linewidth]{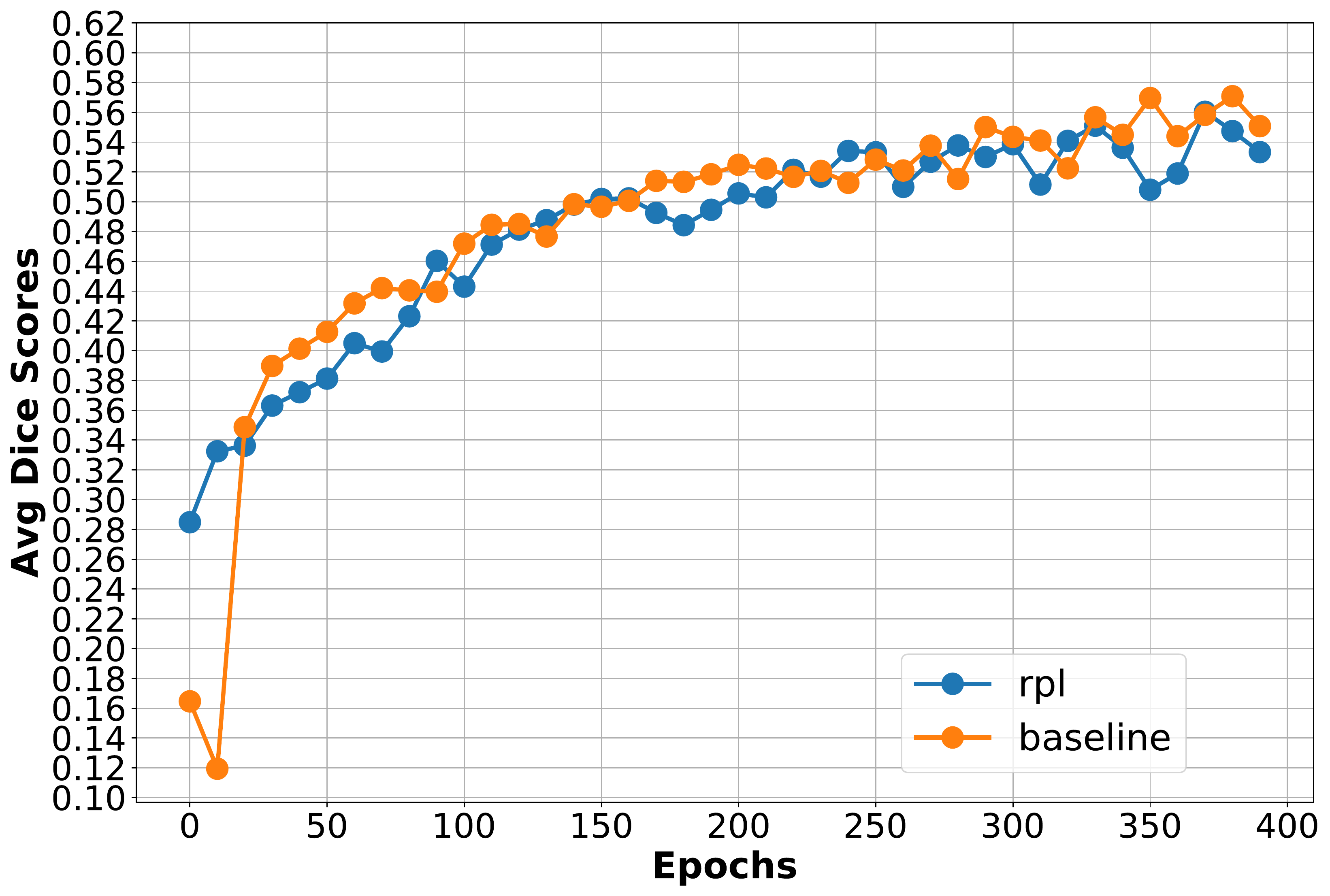}
         \caption{RPL 3D vs. baseline}
     \end{subfigure}
     \hfill
     \begin{subfigure}[b]{0.32\textwidth}
         \centering
         \includegraphics[width=\linewidth]{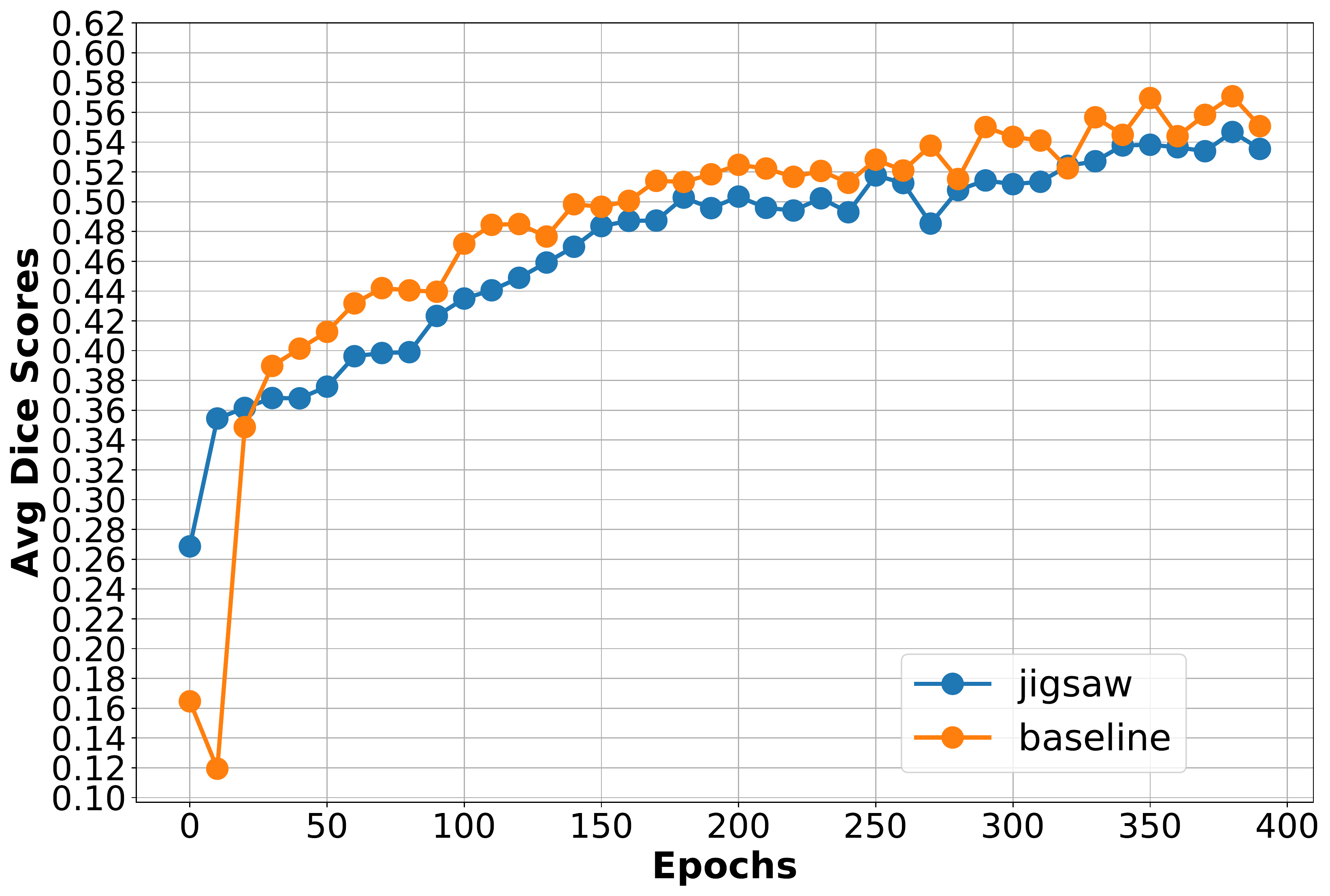}
         \caption{Jigsaw 3D vs. baseline}
     \end{subfigure}
     \vskip\baselineskip
     \begin{subfigure}[b]{0.32\textwidth}
         \centering
         \includegraphics[width=\linewidth]{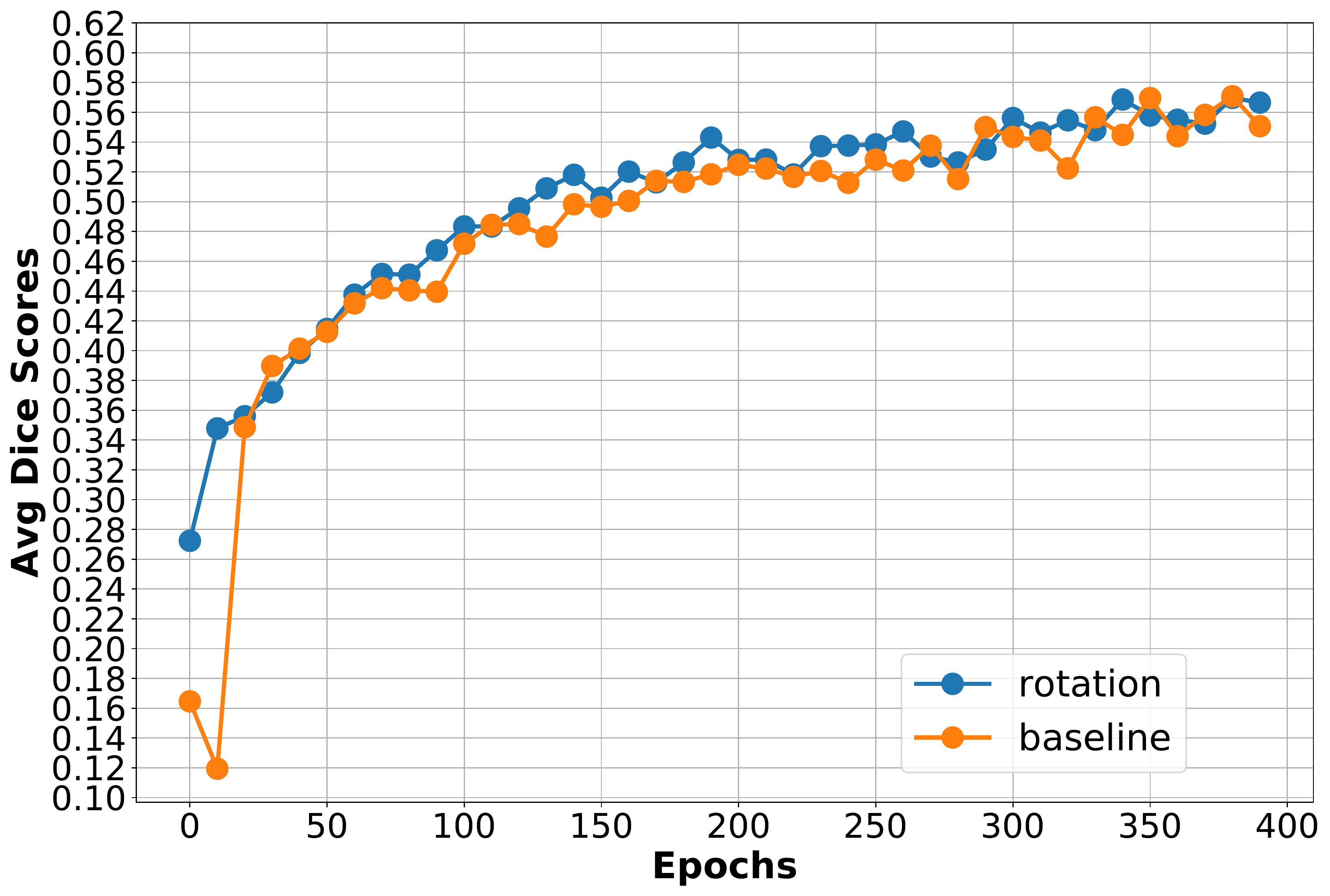}
         \caption{Rotation 3D vs. baseline}
     \end{subfigure}
     \hfill
     \begin{subfigure}[b]{0.32\textwidth}
         \centering
         \includegraphics[width=\linewidth]{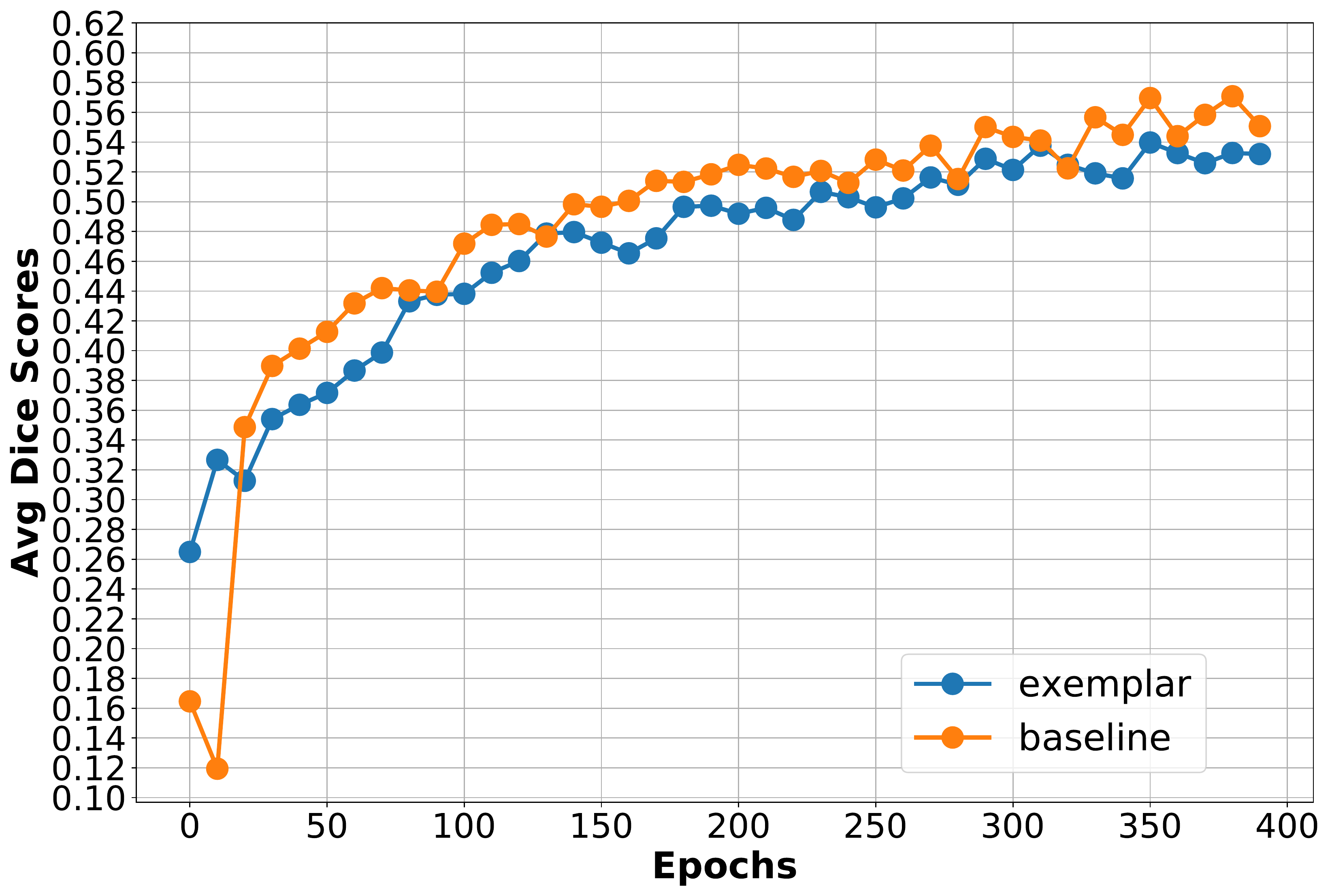}
         \caption{Exemplar 3D vs. baseline}
     \end{subfigure}
    \caption{Pancreas segmentation: Detailed speed of convergence results per method (blue) vs. the supervised baseline (orange). This benefit of our methods helps achieve high results using only few epochs }
\end{figure}

\begin{figure}
    \begin{subfigure}[b]{0.32\textwidth}
        \centering
        \includegraphics[width=\linewidth]{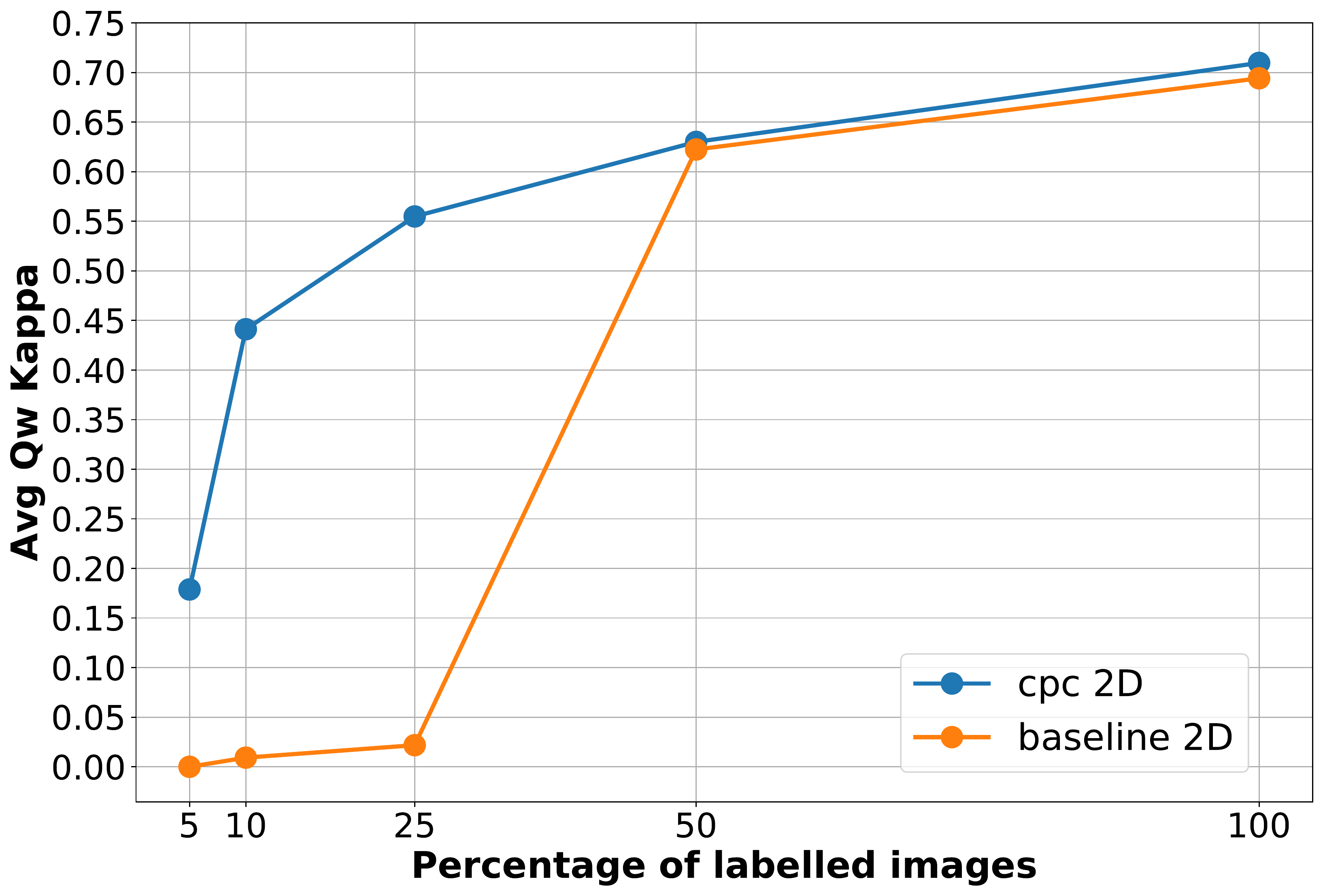}
        \caption{CPC 2D vs. baseline}
    \end{subfigure}
    \hfill
    \begin{subfigure}[b]{0.32\textwidth}
        \centering
        \includegraphics[width=\linewidth]{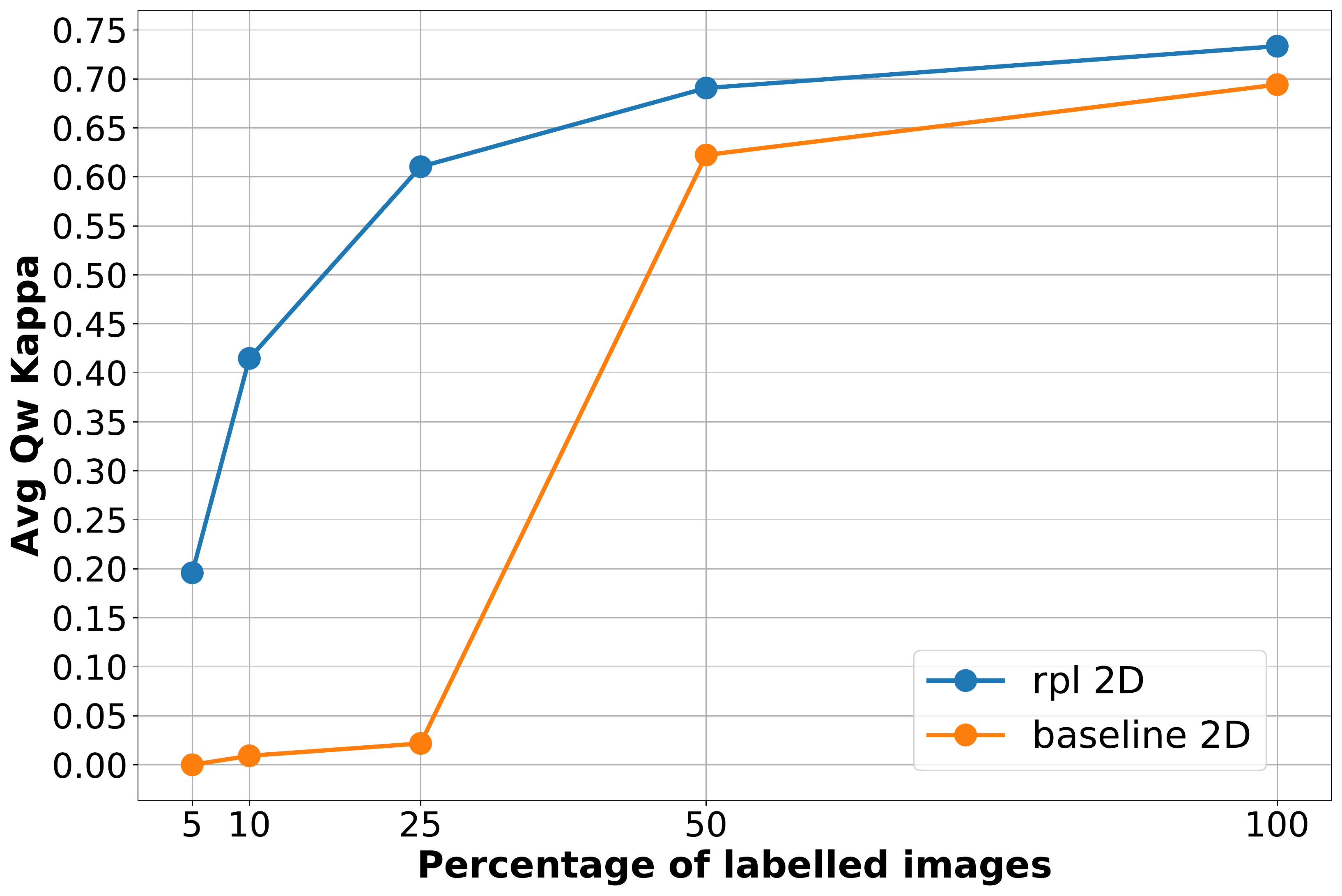}
        \caption{RPL 2D vs. baseline}
    \end{subfigure}
    \hfill
    \begin{subfigure}[b]{0.32\textwidth}
        \centering
        \includegraphics[width=\linewidth]{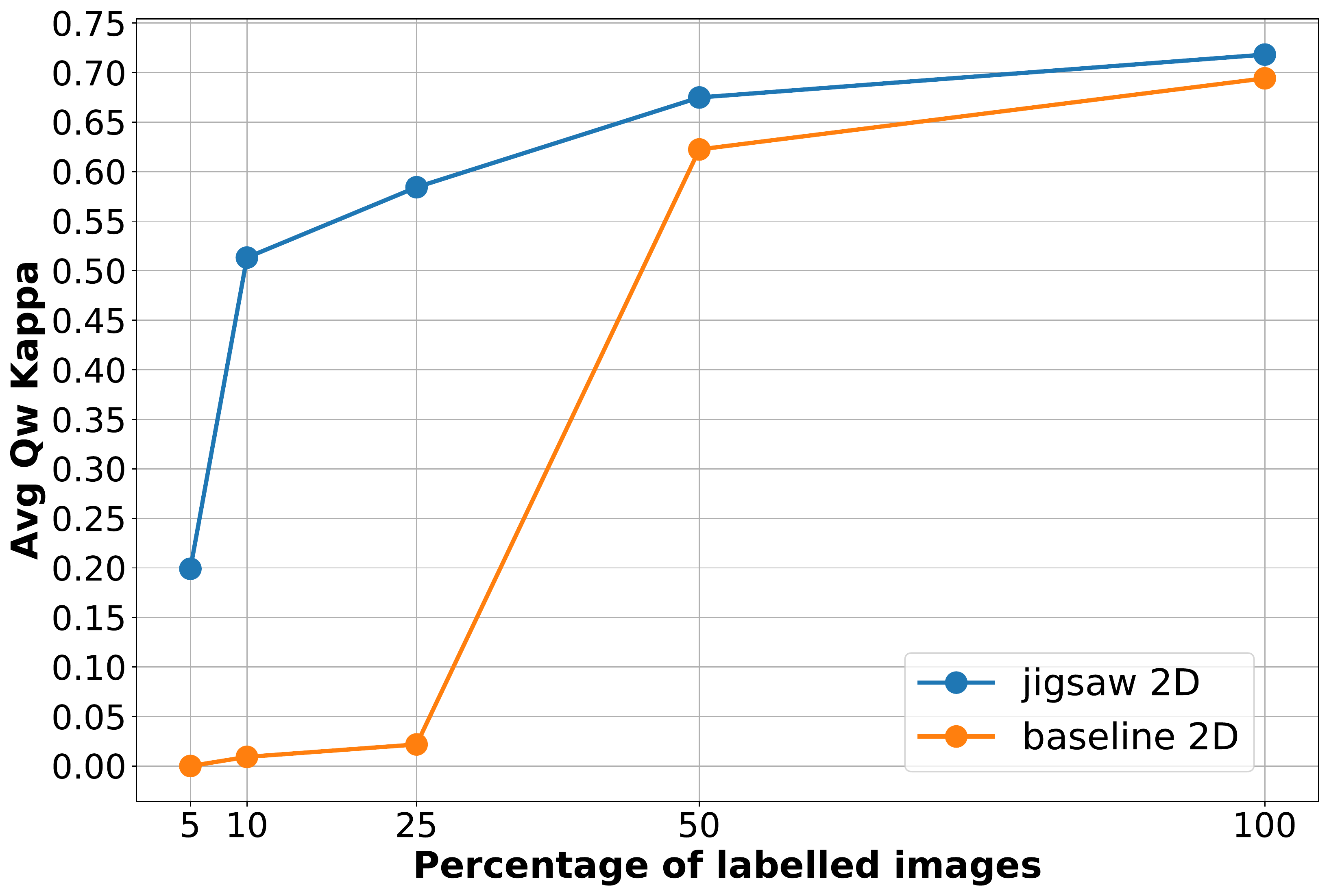}
        \caption{Jigsaw 2D vs. baseline}
    \end{subfigure}
    \vskip\baselineskip
    \begin{subfigure}[b]{0.32\textwidth}
        \centering
        \includegraphics[width=\linewidth]{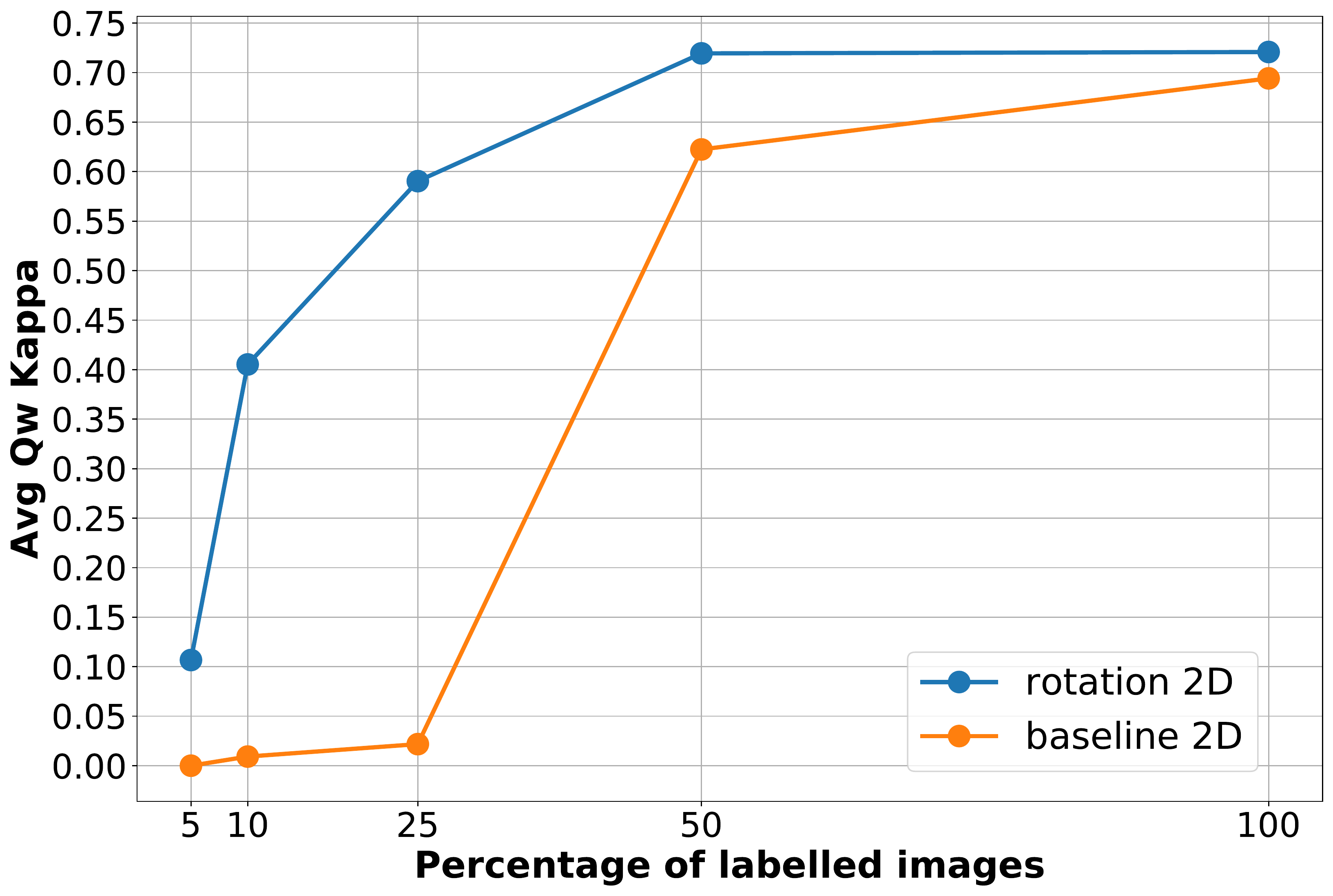}
        \caption{Rotation 2D vs. baseline}
    \end{subfigure}
    \hfill
    \begin{subfigure}[b]{0.32\textwidth}
        \centering
        \includegraphics[width=\linewidth]{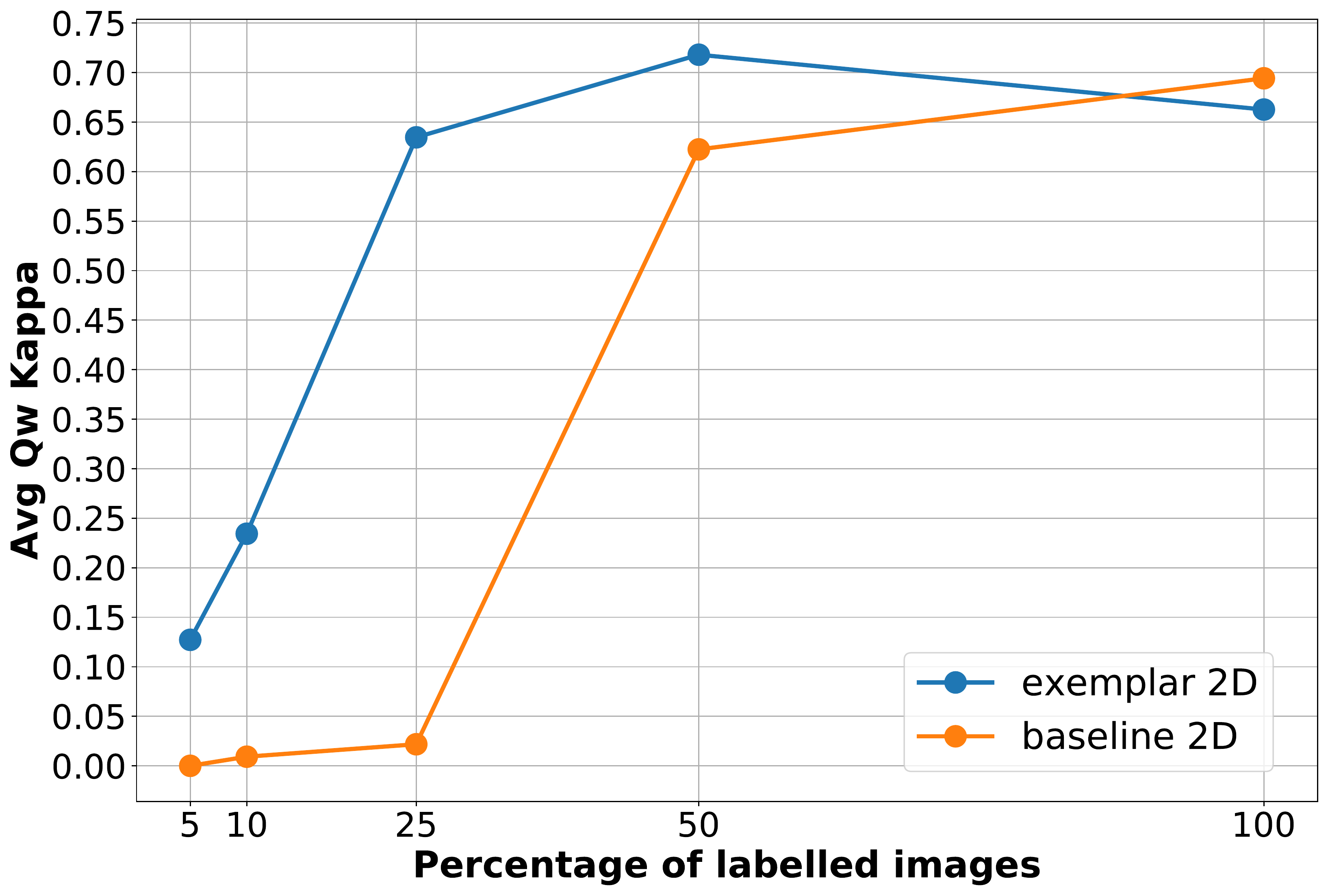}
        \caption{Exemplar 2D vs. baseline}
    \end{subfigure}
    \caption{Retinopathy detection: Detailed data-efficiency results per method (blue) vs. the supervised baseline (orange). Our methods consistently outperform the baseline in low-data cases}
\end{figure}

\begin{figure}
     \begin{subfigure}[b]{0.32\textwidth}
         \centering
         \includegraphics[width=\linewidth]{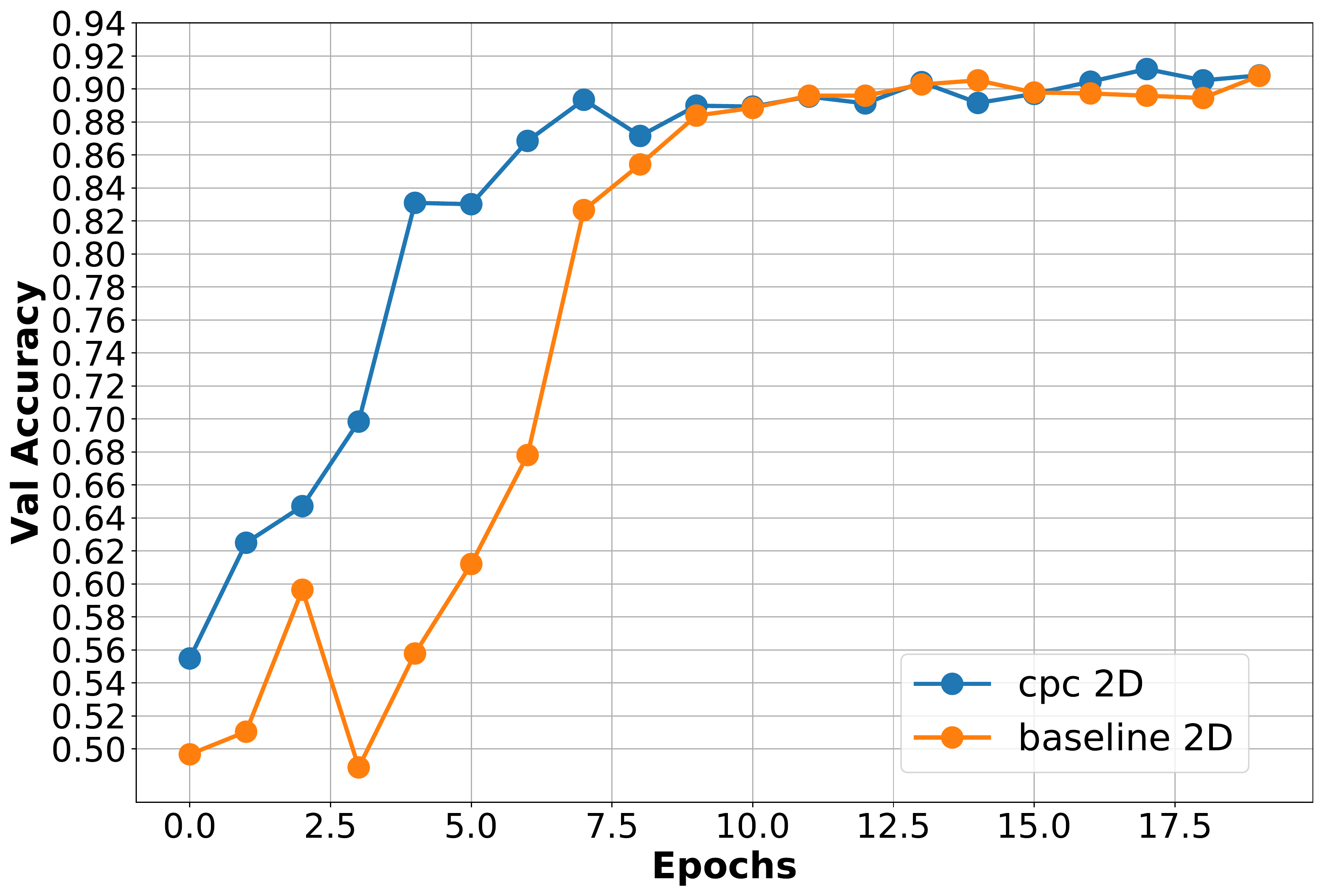}
         \caption{CPC 2D vs. baseline}
     \end{subfigure}
     \hfill
     \begin{subfigure}[b]{0.32\textwidth}
         \centering
         \includegraphics[width=\linewidth]{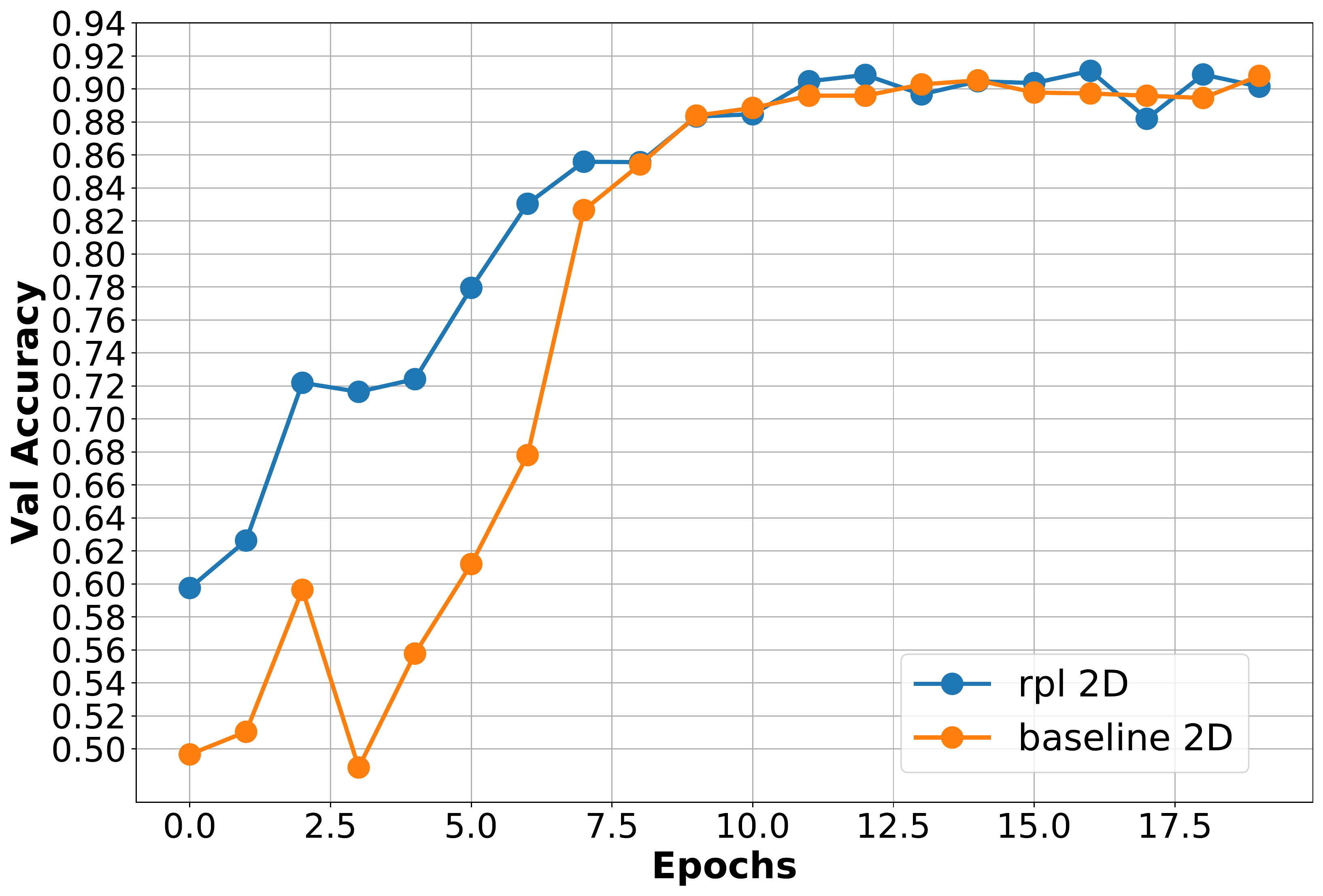}
         \caption{RPL 2D vs. baseline}
     \end{subfigure}
     \hfill
     \begin{subfigure}[b]{0.32\textwidth}
         \centering
         \includegraphics[width=\linewidth]{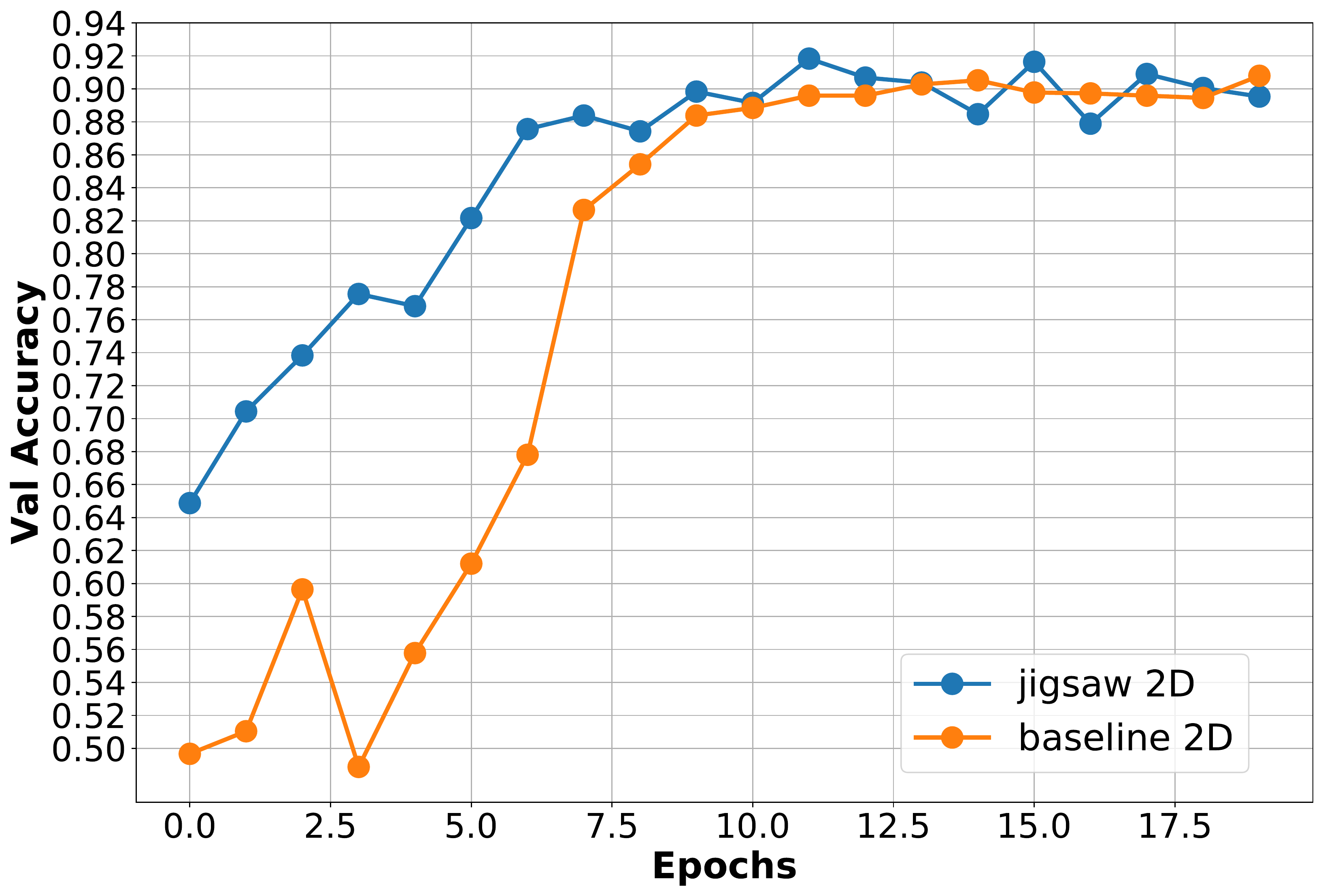}
         \caption{Jigsaw 2D vs. baseline}
     \end{subfigure}
     \vskip\baselineskip
     \begin{subfigure}[b]{0.32\textwidth}
         \centering
         \includegraphics[width=\linewidth]{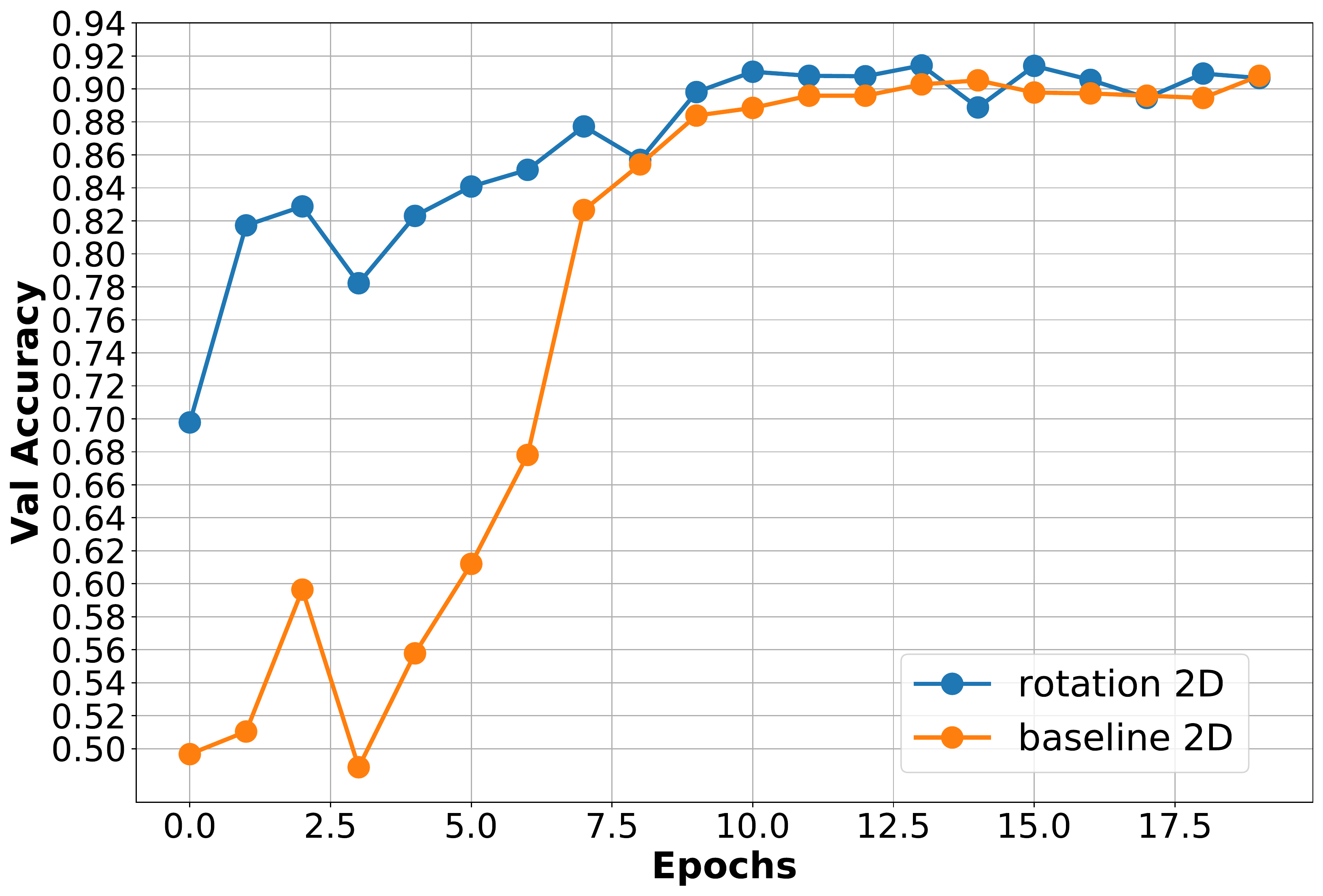}
         \caption{Rotation 2D vs. baseline}
     \end{subfigure}
     \hfill
     \begin{subfigure}[b]{0.32\textwidth}
         \centering
         \includegraphics[width=\linewidth]{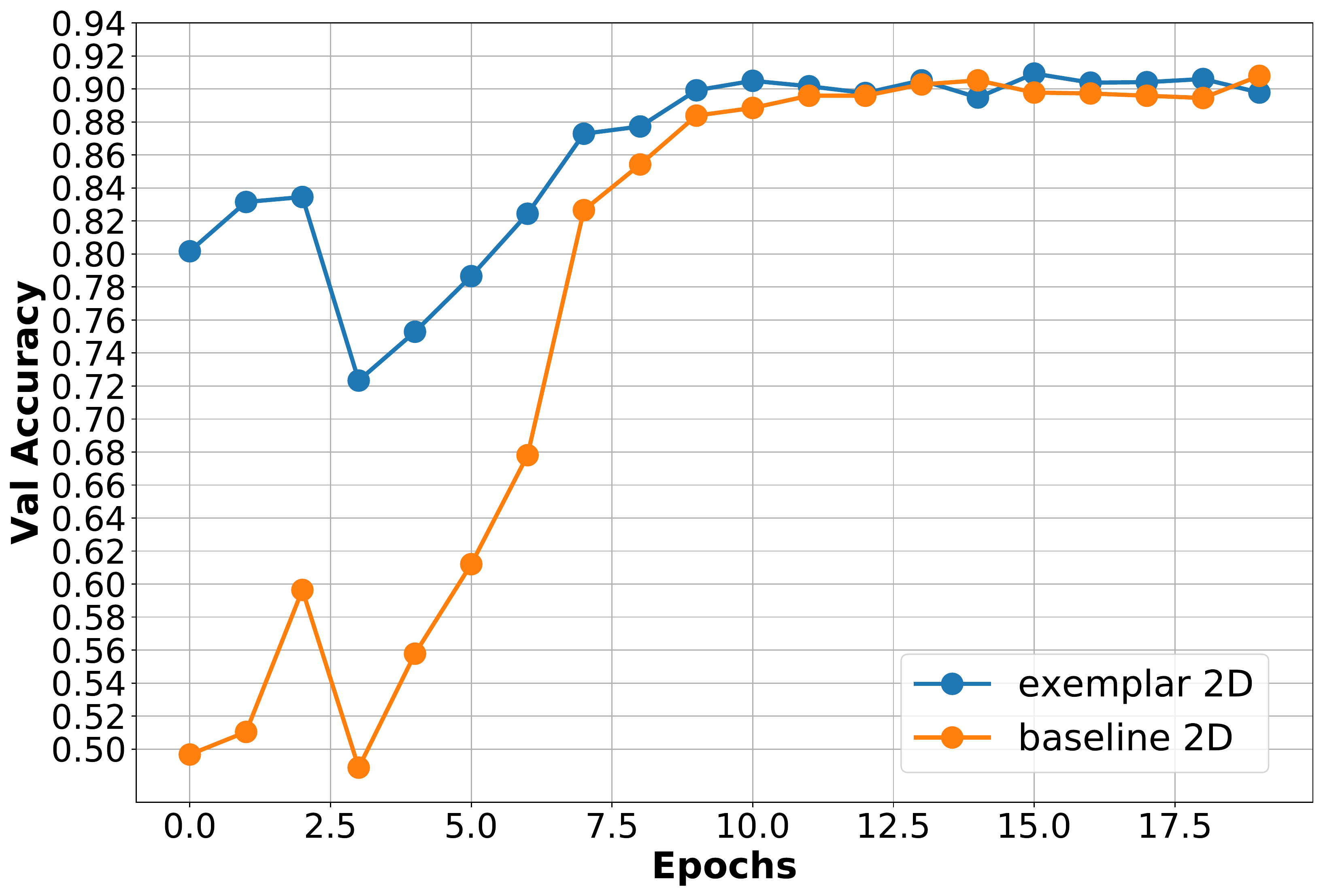}
         \caption{Exemplar 2D vs. baseline}
     \end{subfigure}
    \caption{Retinopathy detection: Detailed speed of convergence results per method (blue) vs. the supervised baseline (orange). This benefit of our methods helps achieve high results using only few epochs }
\end{figure}

\end{appendices}

\end{document}